\pdfoutput=1

\documentclass[11pt]{article}

\usepackage{ACL2023}
\usepackage{times}
\usepackage{latexsym}

\usepackage[T1]{fontenc}

\usepackage[utf8]{inputenc}

\usepackage{microtype}

\usepackage{inconsolata}
\usepackage{textcomp}
\usepackage{amssymb}

\usepackage{adjustbox}
\usepackage{amsmath}
\usepackage{amssymb}
\usepackage{graphicx}
\usepackage{tabularx}
\usepackage{algorithm}
\usepackage{dsfont}
\usepackage{times}
\usepackage{latexsym}
\usepackage{inconsolata}
\usepackage{mathrsfs}
\usepackage{multirow}
\usepackage{tabularx}
\usepackage{color, colortbl}
\usepackage{caption}
\usepackage{booktabs}
\usepackage{subcaption}
\usepackage{mwe}
\usepackage{soul}
\usepackage{xurl}

\usepackage{amsfonts}
\usepackage{bm}
\usepackage{euflag}
\usepackage{tcolorbox}
\usepackage[noend]{algpseudocode}
\definecolor{Gray}{gray}{0.92}
\definecolor{LightCyan}{rgb}{0.92,0.968,0.968}
\usepackage{todonotes}
\newcommand{\notcheckmark}{{$\checkmark$}\textsuperscript{\textcolor{black}{\kern-0.65em{\bf--}}}}

\usepackage{CJKutf8}
\newcommand{\zh}[1]{\begin{CJK}{UTF8}{gbsn}#1\end{CJK}}

\usepackage{xcolor}
\definecolor{myred}{rgb}{0.6, 0.15, 0.3}
\usepackage{todonotes}
\makeatletter
\newcommand*\iftodonotes{\if@todonotes@disabled\expandafter\@secondoftwo\else\expandafter\@firstoftwo\fi}
\makeatother

\definecolor{edolime}{rgb}{0.9,1,0.3}

\newcommand{\rparagraph}[1]{\vspace{1.5mm}\noindent\textbf{#1.}}
\newcommand{\rparagraphnodot}[1]{\vspace{1.5mm}\noindent\textbf{#1}}
\newcommand{\sparagraph}[1]{\vspace{0.0mm}\noindent\textbf{#1.}}

\newcommand{\method}{{\textsc{EnsAd}}\xspace}

\newcolumntype{Y}{>{\centering\arraybackslash}X}

\title{Translation-Enhanced Multilingual Text-to-Image Generation}

\author{Yaoyiran Li$^{\spadesuit,}$\thanks{$^*$This work has been done during the author’s internship
at Amazon Alexa AI.} \quad Ching-Yun Chang$^{\diamondsuit}$ \quad Stephen Rawls$^{\diamondsuit}$ \\
{\bf Ivan Vuli\'{c}}$^{\spadesuit}$ \quad {\bf Anna Korhonen}$^{\spadesuit}$ \\
  $^{\spadesuit}$Language Technology Lab, TAL, University of Cambridge \\
  $^{\diamondsuit}$Amazon Alexa AI\\ \texttt{yl711@cam.ac.uk}, \texttt{\{cychang,sterawls\}@amazon.com} \\ \texttt{\{iv250,alk23\}@cam.ac.uk}\\} 
  
\begin{document}
\maketitle
\begin{abstract}
Research on text-to-image generation (TTI) still predominantly focuses on the English language due to the lack of annotated image-caption data in other languages; in the long run, this might widen inequitable access to TTI technology. In this work, we thus investigate multilingual TTI (termed \textit{mTTI}) and the current potential of neural machine translation (NMT) to bootstrap mTTI systems. We provide two key contributions. \textbf{1)} Relying on a multilingual multi-modal encoder, we provide a systematic empirical study of standard methods used in cross-lingual NLP when applied to mTTI: \textsc{Translate Train}, \textsc{Translate Test}, and \textsc{Zero-Shot Transfer}. \textbf{2)} We propose Ensemble Adapter (\method), a novel parameter-efficient approach that learns to weigh and consolidate the multilingual text knowledge within the mTTI framework, mitigating the language gap and thus improving mTTI performance. Our evaluations on standard mTTI datasets COCO-CN, Multi30K Task2, and LAION-5B demonstrate the potential of translation-enhanced mTTI systems and also validate the benefits of the proposed \method which derives consistent gains across all datasets. Further investigations on model variants, ablation studies, and qualitative analyses provide additional insights on the inner workings of the proposed mTTI approaches.
\end{abstract}


\section{Introduction and Motivation}
\label{s:introduction}

Text-to-Image Generation (TTI) is an emerging yet rapidly growing area, owing its recent progress to ever-growing deep generative models, larger-scale multi-modal datasets, and increasing computational resources. The success of recent TTI work is impressive; e.g., it is possible to synthesise not only high-resolution complex scenes~\cite{Ramesh2022HierarchicalTI,Rombach_2022_CVPR}, but also surrealist and `aesthetics-aware' paintings~\cite{Gallego2022PersonalizingTG}. 

However, current models are made and deployed almost exclusively for the English language (\textsc{en}). This is primarily due to the lack of annotated image-caption data in other languages, which might result in inequitable access to TTI technology in the long run, especially for low-resource languages \cite{blasi-etal-2022-systematic}. Hiring human annotators to write high-quality image descriptions is time-consuming and expensive; `gold standard' data, if it exists at all, is thus typically used for evaluation purposes only \cite{10.1145/3123266.3123366,Aggarwal2020TowardsZC}. 

Even if we put the crucial concerns of data scarcity aside, training state-of-the-art (SotA) TTI models from scratch for each language is technically infeasible and impractical: it would consume massive computational resources, exceeding the capabilities of many research labs~\cite{pmlr-v139-ramesh21a,Saharia2022PhotorealisticTD} and raising concerns of its environmental impact \cite{Schwartz:2020green}.\footnote{For instance, DALL-E~\cite{pmlr-v139-ramesh21a} is trained on $1,024$ $\times$ $16$GB NVIDIA$^{\tiny\circledR}$ V100 GPUs for a total of 430,000 updates. \href{https://huggingface.co/dalle-mini/dalle-mega}{DALL-E Mega}, an attempt to reproduce DALL-E's results, reports an estimated emission of $18,013.47$-kg CO$_{2}$-equivalents, training on a TPU v3-256 ($128$$\times$TPU v3 chips) for $56$ days. The estimation is based on a publicly available machine learning emissions calculator \cite{luccioni2019quantifying}.} Therefore, in this work, we focus on multilingual TTI (\textit{mTTI}) through the optics of NLP's cross-lingual transfer learning methods, leaning on the reasonable assumption of having abundant image-text pairs in English (and/or a pretrained \textsc{en} TTI model), but only limited gold-standard data for fine-tuning and evaluation in a target language.\footnote{A more detailed discussion on data sources, data availability and scarcity is provided in Appendix~\ref{appendix:data}.} 


In particular, we investigate the role of cross-lingual transfer and (neural) machine translation (MT) in bootstrapping mTTI, and we focus on two crucial research questions. \textbf{(RQ1)} Are standard MT-based cross-lingual transfer methods feasible for mTTI, and how do they compare with standard zero-shot cross-lingual transfer methods? \textbf{(RQ2)} Is it possible to enhance zero-shot cross-lingual transfer relying on (ensembles of) MT-generated output for improved mTTI? 

Our experiments and core findings are based on several mTTI benchmarks. First, we use the standard and publicly available COCO-CN~\cite{Li2019COCOCNFC} and Multi30K~\cite{elliott-etal-2016-multi30k}, and we also build a new dataset for Finnish as a lower-resource language from LAION-5B~\cite{Schuhmann2022LAION5BAO}. Regarding RQ1, we then conduct a systematic empirical study comparing the standard cross-lingual transfer methods: \textsc{Translate Train}, \textsc{Translate Test}, and \textsc{Zero-Shot Transfer}. Our main results indicate that \textsc{Translate Train} achieves the best performance, followed by \textsc{Zero-Shot Transfer} which outperforms \textsc{Translate Test}. 

Regarding RQ2, we aim to combine MT-based and zero-shot cross-lingual transfer via fast and parameter-efficient fine-tuning. Inspired by the speech processing literature where a list of Automatic Speech Recognition (ASR) hypotheses can be \textit{jointly considered} for downstream tasks~\cite{ganesan-etal-2021-n,9414806} to alleviate the misrecognition of ASR systems, we propose a module within our mTTI framework termed Ensemble Adapter (\method). It fuses the text encodings of `non-English' text input and a set of its translations to English. Additionally inspired by \citet{Ponti2021ModellingLT}, the idea is to combine the knowledge from \textit{multiple translations} to mitigate potential translation errors, and that way boost cross-lingual transfer for mTTI. 

Our proposed method derives robust gains across all evaluation datasets. Besides offering SotA mTTI performance, the introduced \method component also adds only $0.1$\% dedicated extra parameters (relative to the full mTTI model size) per each supported target language. Put simply, the use of \method increases the portability of our mTTI framework through quick and parameter-efficient adaptation to new languages. The resources of our work are available at \url{https://www.amazon.science/code-and-datasets/translation-enhanced-multilingual-text-to-image-generation}.

\section{Related Work}
\label{s:related_work}

\sparagraph{Text-to-Image Generation} There are generally two categories of standard TTI setups: \textbf{1)} a supervised setup, where gold standard training and test data are from the same domain (e.g., both from MS-COCO); and \textbf{2)} a zero-shot setup, where there is a domain difference between the training data (typically large-scale noisy Web-crawled data) and the high-quality test data (typically manually constructed). GAN-based models are common in supervised TTI setups~\cite{pmlr-v48-reed16,Xu_2018_CVPR,Zhu_2019_CVPR}: they still hold the SotA results, offering smaller model sizes and faster image generation speed~\cite{Zhang_2021_CVPR,Tao_2022_CVPR,Zhou_2022_CVPR}. GigaGAN~\cite{kang2023gigagan}, a recent attempt to scale up GAN models, achieves fairly strong and competitive zero-shot TTI performance. However, in the zero-shot setup, large Vector Quantised Variational Autoencoder (VQVAE)-based models ~\cite{pmlr-v139-ramesh21a,10.1007/978-3-031-19836-6_6,Gafni2022MakeASceneST} and large diffusion models ~\cite{pmlr-v162-nichol22a,Ramesh2022HierarchicalTI,Saharia2022PhotorealisticTD} play the leading role and offer the best performance.

\rparagraph{Multilingual and Non-\textsc{en} TTI} Research on mTTI and non-\textsc{en} TTI is currently limited and only in its infancy. Cogview is a large VQVAE-based Chinese TTI model with training data partly from crawling Chinese websites and social media platforms, and partly from translating \textsc{en} data~\cite{NEURIPS2021_a4d92e2c}. ruDALL-E is a VQVAE-based Russian TTI model recreating DALL-E \cite{pmlr-v139-ramesh21a} with training data translated from \textsc{en} data.\footnote{\href{https://rudalle.ru/}{https://rudalle.ru/}; ruDALL-E has not released an accompanying paper yet, but a \href{https://habr.com/ru/company/sberbank/blog/589673/}{technical blog} is available.}


To the best of our knowledge, there are only two existing papers attempting multilingual or cross-lingual TTI. \newcite{ZHANG2022108006} align two monolingual text encoders, one for the source and the other for the target language, with a fixed image generator pretrained on the source language (i.e., \textsc{en}). \newcite{jung-etal-2022-language} take a step further, relying on a multilingual text encoder that supports more languages simultaneously. 

We note several crucial differences to the prior work. \textbf{1)} The two papers are based on earlier TTI models~\cite{Xu_2018_CVPR}, which are now largely surpassed by recent SotA models~\cite{Zhou_2022_CVPR}. \textbf{2)} Their model designs are tied to the model of ~\newcite{Xu_2018_CVPR} and cannot be easily adapted to the latest SotA TTI models. \textbf{3)} They use traditional LSTM text encoders enhanced by mono-modal BERT features, while SotA TTI models~\cite{Zhou_2022_CVPR,Saharia2022PhotorealisticTD,Rombach_2022_CVPR} use the multi-modal CLIP model~\cite{pmlr-v139-radford21a}. Therefore, we neither adopt them as baselines nor try to adapt them for our use, also taking into account the difficulty of replicating the prior work as no code has been released to date. In contrast, our work relies on the mCLIP text encoder~\cite{carlsson-etal-2022-cross}, the multilingual version of CLIP, and is developed based on LAFITE~\cite{Zhou_2022_CVPR}, a SotA TTI model. In fact, as shown later in our work, training an English TTI model using mCLIP without any further tuning can \textit{already realise} zero-shot mTTI, similar to what has been attempted by~\newcite{jung-etal-2022-language}.



\rparagraph{Translation-Based Cross-lingual Transfer} Machine translation (MT) at both lexical level and sentence level has been successfully used for cross-lingual transfer learning in NLP, where \textsc{Translate Train} and \textsc{Translate Test} usually serve as strong baselines for downstream tasks~\cite{conneau-etal-2018-xnli,glavas-etal-2019-properly,pmlr-v119-hu20b,Ponti2021ModellingLT,li-etal-2022-improving,li-etal-2022-improving-bilingual}. In addition, MT is used to generate sentence pairs for training multilingual multi-modal models~\cite{Zhou_2021_CVPR,carlsson-etal-2022-cross}. However, MT is still largely underexplored and underutilised for mTTI. In this work, we analyse the potential of MT to enhance multilingual and cross-lingual TTI.

\section{Methodology}
\label{s:methodology}
In what follows in this section, we first introduce our base mLAFITE model and three baseline approaches for mTTI (\S\ref{s:methodology_mlafite}). Next, we propose an Ensemble Adapter module that can work in synergy with the pretrained mLAFITE model to improve mTTI performance (\S\ref{s:methodology_ensad}). Finally, we describe how we train our Ensemble Adapter and formulate our loss functions (\S\ref{s:methodology_training}).

\subsection{mLAFITE and Baselines}
\label{s:methodology_mlafite}
For easier deployment and comparison of different cross-lingual transfer methods, our work focuses on the relatively lightweight GAN-based models, which are faster to train and evaluate compared with VQVAE-based models and large diffusion models (see \S\ref{s:related_work}). In particular, we adopt LAFITE~\cite{Zhou_2022_CVPR}, a SotA GAN-based English TTI model, as our starting point. To unlock its multilingual capabilities, we replace its English-only CLIP text encoder~\cite{pmlr-v139-radford21a} with mCLIP~\cite{carlsson-etal-2022-cross}, which is already pretrained to align the sentence representation spaces of $68$ languages.\footnote{mCLIP is derived by fine-tuning a pretrained XLM-R model \cite{carlsson-etal-2022-cross,conneau-etal-2020-unsupervised}, and it does not directly depend on parallel corpora or multilingual image-text data. The work uses NMT to generate `silver'-quality \textsc{en}-$*$ sentence pairs and then directly aligns the CLIP-extracted \textsc{en} representations and mCLIP's multilingual sentence representations of the NMT-generated data. Both CLIP and mCLIP use a shared CLIP visual encoder.} 

There are three common categories of cross-lingual transfer approaches which we apply to mTTI and adopt as our principal baselines: 

\vspace{1.5mm}
\noindent \textbf{\textsc{Translate Train}}. We translate all the captions from the English training set (e.g., COCO) into a (non-\textsc{en}) target language ($\mathbb{L}$) relying on an MT system. We then train a LAFITE TTI model in the target language from scratch, relying on mCLIP as the text encoder.\footnote{We use mCLIP rather than monolingual CLIP since it is infeasible for most languages. Only several high-resource languages have publicly available monolingual models. For fair cross-language comparisons, we leverage the same mCLIP text encoder in all our experiments.} At inference, an $\mathbb{L}$ sentence is directly fed into the target-language TTI model. 

\vspace{1.5mm}
The other two approaches instead rely on a TTI model pretrained with \textit{English} data, and they do \textit{not} require further tuning with captions in the target languages. As our first step, we pretrain an mCLIP-based LAFITE model (we call it \textbf{mLAFITE} for brevity) from scratch.   

\vspace{1.5mm}
\noindent \textbf{\textsc{Translate Test}}. At inference, we first translate a caption in $\mathbb{L}$ into \textsc{en} via MT and the \textsc{en} translation then serves as mLAFITE's input.

\vspace{1.5mm}
\noindent \textbf{\textsc{Zero-Shot Transfer}}. Since mCLIP is a multilingual sentence encoder, text in $\mathbb{L}$ can be directly fed to our mLAFITE for TTI without any extra fine-tuning.

\subsection{mLAFITE with Ensemble Adapter}
\label{s:methodology_ensad}
\begin{figure*}[t!]
\centering
\includegraphics[width=0.86\linewidth]{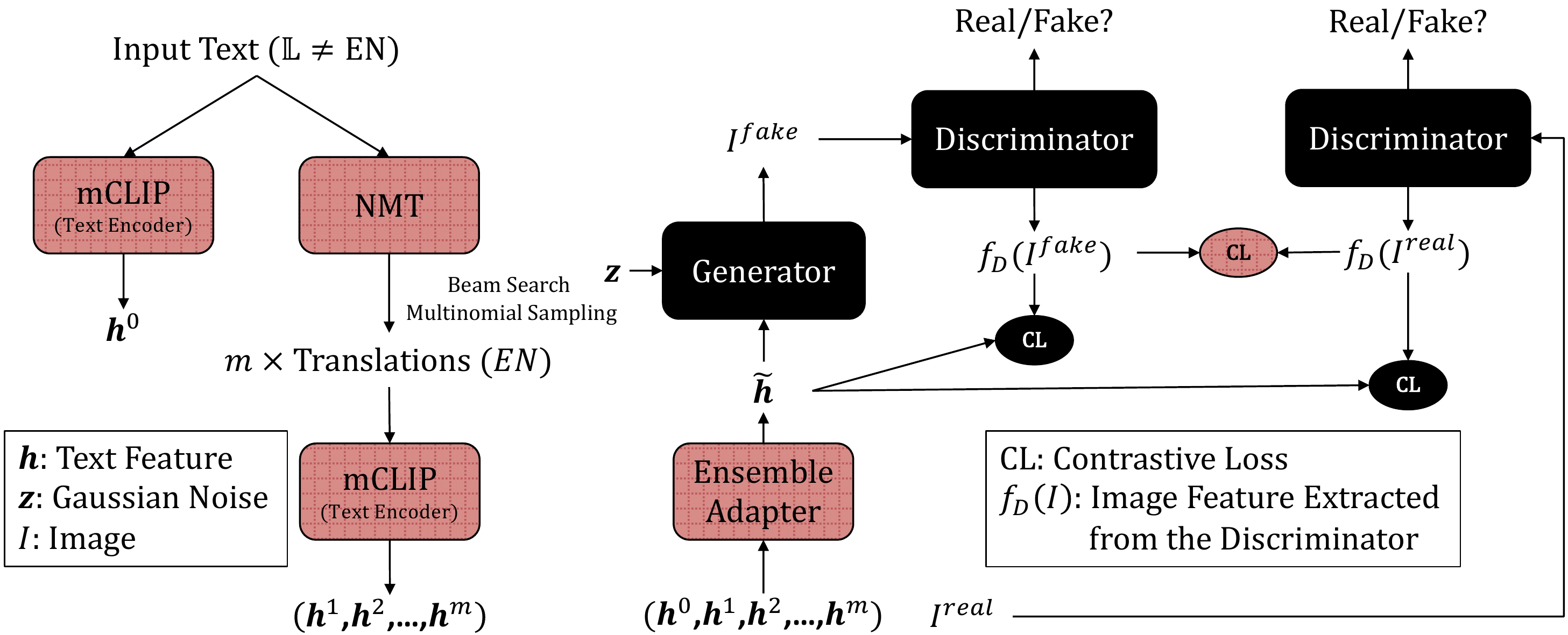}
\caption{An overview of the full proposed mTTI framework with the Ensemble Adapter module. The black blocks are networks and contrastive learning (CL) losses already in the original LAFITE model (also in our pretrained mLAFITE). Our proposed, newly added modules, and a CL loss are provided in red, gridded blocks.}
\label{fig:arch}
\end{figure*}

We now propose an attention-based Ensemble Adapter (\method) module that aims to improve mTTI via leveraging knowledge from multiple translations of the same input. The full pipeline and how \method extends the base mLAFITE model are illustrated in Figure~\ref{fig:arch}.
Given an input sentence in language $\mathbb{L}$, $\mathbb{L}$$\neq$\textsc{en}, we first use any (N)MT system to sample a set of \textsc{en} translations. We then deploy the \method module between the mCLIP text encoder and the TTI generator to fuse the mCLIP-extracted embeddings, bridging the \textsc{en}-$\mathbb{L}$ language domain gap. The adapter can be trained with only a small set of image-$\mathbb{L}$ text pairs while mCLIP and the TTI generator networks are kept frozen.


Formally, we use $x^{0}$ to denote the $\mathbb{L}$ input text, while $\{x^{1},x^{2},...,x^{m}\}$ is a set of $m$ \textsc{en} translations of the $\mathbb{L}$ input text. The fixed mCLIP encoder extracts their respective ($l_2$-normalised) $d$-dimensional sentence embeddings, yielding the matrix $\bm{H}=(\mathbf{h}^{0},\mathbf{h}^{1},...,\mathbf{h}^{m})\in\mathbb{R}^{d \times (m+1)}$. Then, our proposed \method learns to fuse these sentence encodings from $\bm{H}$. We define the query ($\mathbf{q}$), key ($\bm{K}$), and value ($\bm{V}$) inputs of our attention as:
%
\begin{align}
\begin{split}\label{formula:Attn_QKV1}
\mathbf{q}&=\mathbf{h}^{0},
\end{split}\\
\begin{split}\label{formula:Attn_QKV2}
\bm{K}&=(\mathbf{h}^{1},\mathbf{h}^{2},...,\mathbf{h}^{m}),
\end{split}\\
\begin{split}\label{formula:Attn_QKV3}
\bm{V}&=(\mathbf{h}^{1}-\mathbf{h}^{0},\mathbf{h}^{2}-\mathbf{h}^{0},...,\mathbf{h}^{m}-\mathbf{h}^{0}).
\end{split}
\end{align}
Note that $\{\mathbf{h}^{0},\mathbf{h}^{1},...,\mathbf{h}^{m}\}$ are all close to each other in the mCLIP representation space. Therefore, to focus on the `additional information' contained in the \textsc{en} translations,  we take the difference between $\mathbf{h}^{i},i>0$ and $\mathbf{h}^{0}$ as in Eq.~\eqref{formula:Attn_QKV3}.\footnote{We adopt the simple mean pooling of $\{\mathbf{h}^{0},\mathbf{h}^{1},...,\mathbf{h}^{m}\}$ as an additional baseline with results in \S\ref{s:RQ2results}. We also tried multi-head self-attention~\cite{NIPS2017_3f5ee243}, where $\bm{Q}=\bm{K}=\bm{V}=\bm{H}$, which, however, showed inferior performance in our preliminary experiments.} The calculation of attention scores is then based on the standard additive attention~\cite{Bahdanau2015NeuralMT}:
%
\begin{align}
\begin{split}\label{formula:Attn_Scores1}
\bm{A}&=\bm{W}^{q}\mathbf{q}\mathds{1}^{\text{T}}+\bm{W}^{k}\bm{K}+\bm{W}^{v}\bm{V}+\mathbf{b}\mathds{1}^{\text{T}},
\end{split}\\ 
\begin{split}\label{formula:Attn_Scores2}
\mathbf{s}^{\text{T}}&=\text{softmax}(\bm{W}^{p}(\text{tanh}(\bm{A}))).
\end{split}
\end{align}
\noindent \method's hidden size is $d_{hid}$; $\bm{W}^{q},\bm{W}^{k},\bm{W}^{v}\in\mathbb{R}^{d_{hid}\times d}$ are respective mappings for query, key, and value inputs; $\mathbf{b}\in\mathbb{R}^{d_{hid}}$ is the bias, and $\bm{W}^{p}\in\mathbb{R}^{1\times d_{hid}}$ is a final projection matrix for deriving the attention scores. Then, the context vector is an attention-guided summarisation of $\bm{V}$. \method's final output is the linear combination of $\mathbf{h}_{0}$ and the context vector, computed as follows:
%
\begin{align}
\begin{split}\label{formula:Attn_Output1}
\bm{V}^{o}&=(1-\alpha)\bm{V}+\alpha\cdot\text{tanh}(\bm{W}^{o}\bm{V}),
\end{split}\\ 
\begin{split}\label{formula:Attn_Output2}
\mathbf{c}&=\bm{V}^{o}\mathbf{s},
\end{split}\\ 
\begin{split}\label{formula:Attn_Output3}
\tilde{\mathbf{h}}&=\method(\bm{H})=(1-\alpha)\mathbf{q}+\alpha\cdot\mathbf{c},
\end{split}\
\end{align}
\noindent where $\bm{W}^{o}\in\mathbb{R}^{d\times d}$ is the output mapping, and $\alpha$ is an interpolation hyperparameter. We also $\mathit{l}_{2}$-normalise the outputs of Eqs.~\eqref{formula:Attn_QKV3}, \eqref{formula:Attn_Output2}, \eqref{formula:Attn_Output3}, as well as the $\text{tanh}(\bm{W}^{o}\bm{V})$ term in Eq.~\eqref{formula:Attn_Output1}.

\subsection{Contrastive Adversarial Training}
\label{s:methodology_training}
Our Generator ($G$) and Discriminator ($D$) network structures and the pretraining process of the base mLAFITE model all follow LAFITE's original implementation for supervised TTI. As illustrated in Figure~\ref{fig:arch}, we take the pretrained mLAFITE and insert the \method between mCLIP and $G$. We then adversarially train \mbox{\method} and $D$ iteratively while mCLIP and $G$ are kept frozen.\footnote{We also tried freezing $D$ but this results in inferior performance in our preliminary investigation.} Additionally, we propose to optimise a novel contrastive objective aligning the $D$-extracted real image and fake (synthesised) image features in adversarial training. 


The (m)LAFITE GAN framework is adapted from the popular unconditional StyleGAN2 framework~\cite{Karras_2020_CVPR} which features a redesigned adaptive instance normalization mechanism~\cite{Huang_2017_ICCV} in $G$: it enables the unconditional channel-wise `style information' (e.g., pose, lighting, background style) to control $G$'s image synthesis backbone (convolution and upsampling layers). The `style information' is derived as follows: a random noise $\mathbf{z}$ is sampled from the standard Gaussian distribution $\mathcal{N}(\mathbf{0},\mathbf{I})$ and transformed into a so-called unconditional \textit{StyleSpace}, which is proven to be a well-disentangled intermediate latent space~\cite{Wu_2021_CVPR}.\footnote{The transformation includes a shared $8$-layer MLP and a dedicated affine mapping per each generation layer. We refer the reader to the original work for further technical details.} LAFITE further proposes to inject text-conditioning information into the StyleSpace via a series of non-linear and affine mappings. In our pipeline, $G$ takes our \method-gathered feature $\tilde{\mathbf{h}}$ and noise $\mathbf{z}$, and it then outputs a fake image: $\mathcal{I}^{fake}=G(\tilde{\mathbf{h}},\mathbf{z})$.

The discriminator has a characteristic `two-branch' design: \textbf{1)} $D$ is in essence a convolutional \textit{image encoder}, producing $f_{D}(\mathcal{I})$, a $d$-dim image feature for any real or fake (i.e., synthesised) input image $\mathcal{I}$; \textbf{2)} $D$ also predicts if $\mathcal{I}$ is real or fake based on both $\mathcal{I}$ and $\tilde{\mathbf{h}}$, where the prediction (a scalar output) is denoted as $D(\mathcal{I},\tilde{\mathbf{h}})=D_{s}(\mathcal{I})+\tilde{\mathbf{h}}^{T}f_{D}(\mathcal{I})$. This is realised via adding two affine transformations on top of a shared visual backbone for deriving $f_{D}(\mathcal{I})$ and $D_{s}(\mathcal{I})$, respectively. We then define the adversarial (AD) losses for \method and $D$ following LAFITE:
%
%
\begin{align}
\begin{split}\label{formula:LG}
\mathcal{L}_{AD}^{\method}&=-\frac{1}{n}\!\sum_{i=1}^{n}\text{log}\sigma(D(\mathcal{I}^{fake}_{i},\tilde{\mathbf{h}}_{i})),
\end{split}\\ 
\begin{split}\label{formula:LD}
\mathcal{L}_{AD}^{D}=&-\frac{1}{n}\!\sum_{i=1}^{n}\text{log}\sigma(D(\mathcal{I}^{real}_{i},\tilde{\mathbf{h}}_{i}))\\&\!-\frac{1}{n}\!\sum_{i=1}^{n}\text{log}(1-\sigma(D(\mathcal{I}^{fake}_{i},\tilde{\mathbf{h}}_{i}))).
\end{split}
\end{align}
%
\noindent $n$ is the batch size, and $\sigma(\cdot)$ is the sigmoid function. We propose an auxiliary contrastive loss, aligning the discriminator-extracted $\mathcal{I}^{fake}$ and $\mathcal{I}^{real}$ features, computed as follows: 
%
\begin{align}
\begin{split}\label{formula:CL-0}
s_{i,j}=&\text{cos}(f_{D}(\mathcal{I}_{i}^{real}),f_{D}(\mathcal{I}_{j}^{fake})),
\end{split}\\ 
\begin{split}\label{formula:CL}
\mathcal{L}_{CL}=&-\frac{1}{n}\!\sum_{i=1}^{n}\text{log} \frac{\text{exp}(s_{i,i}/\tau)}{\sum_{j=1}^{n}\text{exp}(s_{j,i}/\tau)}.
\end{split}
\end{align}
\noindent $\text{cos}(\cdot)$ calculates the cosine similarity, and $\tau$ is the temperature.

In the original LAFITE paper, there are already two auxiliary contrastive losses: \textbf{1)} $\mathcal{L}_{CL}^{G}$ aligns CLIP-extracted image features of $\mathcal{I}^{fake}$ and the input text embedding, i.e., $\tilde{\mathbf{h}}$ in our case; \textbf{2)} $\mathcal{L}_{CL}^{D}$ aligns $f_{D}(\mathcal{I})$ with its associated $\tilde{\mathbf{h}}$.\footnote{As with LAFITE's original implementation, $f_{D}(\mathcal{I})$ is $f_{D}(\mathcal{I}^{fake})$ in $\mathcal{L}_{\method}$ and $f_{D}(\mathcal{I}^{real})$ in $\mathcal{L}_{D}$.} In our preliminary experiments, we found that $\mathcal{L}_{CL}^{G}$ was not useful for \method, so we completely remove it.\footnote{The equations for the other two CL losses are similar to Eq.~\eqref{formula:CL}. For brevity, we skip the details and refer the reader to the original LAFITE paper.} Our final losses for training \method and $D$ are as follows, with two hyperparameters $\lambda_{1}$ and $\lambda_{2}$ controlling the weights of contrastive losses:
%
\begin{align}
\begin{split}\label{formula:Loss-EA}
\mathcal{L}_{\method}=&\mathcal{L}_{AD}^{\method} + \lambda_{1}\cdot\mathcal{L}_{CL} + \lambda_{2}\cdot\mathcal{L}_{CL}^{D},
\end{split}\\ 
\begin{split}\label{formula:Loss-D}
\mathcal{L}_{D}=&\mathcal{L}_{AD}^{D} + \lambda_{1}\cdot\mathcal{L}_{CL} + \lambda_{2}\cdot\mathcal{L}_{CL}^{D}.
\end{split}
\end{align}
\noindent The full training process is also summarised in Algorithm~\ref{alg}, available in Appendix~\ref{app:alg}. Note that the use of \method introduces only up to $0.1$\% extra parameters per each target language relative to the full model size. This parameter efficiency boosts the portability of our mTTI framework, enabling quick and efficient adaptation to new languages.

\section{Datasets}
\label{s:datasets}

mLAFITE pretraining is based on the MS-COCO~\cite{Chen2015MicrosoftCC} training set comprising $82,783$ images, where each image is associated with 5 \textsc{en} captions. $10\%$ of the training set is held out as our dev set,  and the rest is used for training. MS-COCO also provides a validation set ($40,504$ images), frequently used for TTI evaluation.

For mTTI, we choose evaluation datasets that satisfy the following criteria: \textbf{a)} no overlap between images in the test set and images used in pretraining; \textbf{b)} the test set includes at least $5K$ images;\footnote{Previous work proved that small test set sizes result in biases and unreliable TTI evaluation~\cite{Chong_2020_CVPR}; therefore, TTI work typically adopts test sets with more than $5K$ images \cite{Zhou_2022_CVPR,pmlr-v139-ramesh21a}. For instance, the most common \textsc{en} TTI data for evaluation is the MS-COCO validation set that contains $40K$ images. The smallest general-domain test set in \newcite{Zhou_2022_CVPR} is LN-COCO~\cite{10.1007/978-3-030-58558-7_38} containing $\sim5K$ images.} \textbf{c)} the captions are human-written descriptions and not (manual or MT-derived) translations from \textsc{en} captions.\footnote{Human-written descriptions are more realistic for real-world non-\textsc{en} users, and translations from \textsc{en} captions can cause unexpected `translationese' bias~\cite{elliott-etal-2016-multi30k,van-miltenburg-etal-2017-cross,pmlr-v162-bugliarello22a}.} Based on these requirements, we select three `non-\textsc{en}' datasets, outlined in what follows. 


\rparagraphnodot{COCO-CN}~\cite{Li2019COCOCNFC} provides Chinese (\textsc{zh}) captions (i.e., human descriptions) for $20,341$ MS-COCO images. $6,748$ of them are from the COCO validation set not seen during mLAFITE pretraining; we thus use them as our test set. We randomly sample $20\%$ of the rest as our dev set ($2,718$), and the training set has $10,875$ images. Each image has only one \textsc{zh} caption. COCO-CN additionally offers $5,000$ \textsc{zh} sentences manually translated from \textsc{en} captions; we only use the corresponding \textsc{en}-\textsc{zh} sentence pairs to calculate BLEU scores for comparing different MT systems.

\rparagraphnodot{Multi30K Task2}~\cite{elliott-etal-2016-multi30k,elliott-etal-2017-findings} has $5$ German (\textsc{de}) captions (human descriptions) for each of $31,014$ Flickr30K~\cite{young-etal-2014-image} images. We randomly sample and keep one caption per each image.\footnote{All our non-\textsc{en} TTI datasets uniformly have one caption for each image. This setup is also more realistic since real-world users only need to input a single sentence into a TTI model. For mLAFITE pretraining, however, as with all other related work, all the COCO captions are used.} We randomly split the data into train, dev, and test sets spanning $10,000$, $2,000$, and $19,014$ images, respectively.

\rparagraphnodot{LAION-5B}~\cite{Schuhmann2022LAION5BAO} is a large-scale Web-crawled vision-language dataset with $5$ billion image-text pairs covering $100+$ languages.  We focus on Finnish (\textsc{fi}) as a lower-resource language for our evaluation. Unlike carefully annotated COCO-CN and Multi30K, LAION-5B's data are noisy, so we rely on massive filtering to select relatively high-quality data. The full data creation process for \textsc{fi} is provided in Appendix~\ref{app:fi}. 

The final dataset comprises training, development and test portions with $10,000$, $2,000$, and $18,000$ image-text pairs, respectively. Our manual inspection of the final dataset indicates that it is of acceptable quality although having its own characteristics (Appendix~\ref{app:fi}) but the quality in general still cannot match COCO-CN or Multi30K. We use the data in our main experiments \textbf{1)} as an initial trial to extend TTI evaluation to `non-COCO-style' captions and another language and \textbf{2)} for comparative analyses with COCO-CN and Multi30K. 

\rparagraph{Supplementary Dataset: IGLUE} In order to further widen the set of target languages, we also experiment with \textit{IGLUE xFlickr$\&$CO}~\cite{pmlr-v162-bugliarello22a}. It provides $2K$ images, where one half comes from the MS-COCO validation set and the other half from Multi30K with associated human descriptions in $5$ additional languages: Spanish (\textsc{es}), Indonesian (\textsc{id}), Japanese (\textsc{ja}), Russian (\textsc{ru}), and Turkish (\textsc{tr}). Since IGLUE does not offer a training set, we use it only for RQ1-related experiments. Although IGLUE does not comply with our criterion \textbf{b)} above, we use it to extend our empirical analyses to more languages.


Table~\ref{table:languages_details} in Appendix~\ref{appendix:languages} provides a full and systematic overview of languages and data statistics used in this work.

\section{Experimental Setup}
\label{s:experiments}
In what follows, we outline our experimental setups and choices related to the two core RQs from \S\ref{s:introduction}. We also show details concerning our mLAFITE pretraining, side experiments (most are RQ2-related), and evaluation metric.

 \rparagraph{mLAFITE Pretraining} All methods for mTTI are implemented based on our pretrained mLAFITE model, which is trained with $8$$\times$$16$GB V100 GPUs for $75$ hours (i.e., $40$ million data points sampled from the training set). Contrastive loss weights and other hyper-parameters follow the original LAFITE setup~\cite{Zhou_2022_CVPR}.\footnote{The original LAFITE model is based on \textsc{en} CLIP and is trained with $25$ million samples on $4$$\times$$16$GB V100 GPUs. Our mLAFITE model uses mCLIP, which is of the same dimensionality as CLIP and keeps the values of all the other relevant hyper-parameters, such as batch per GPU and learning rate, from the original work.} For fair comparisons, we use the same mCLIP text encoder for all our RQ1 and RQ2 experiments. 
 

\rparagraph{RQ1 Experiments} 
On COCO-CN, we compare four widely used MT systems: Amazon Translate\footnote{\href{https://aws.amazon.com/translate/}{https://aws.amazon.com/translate/}}, a SotA commercial MT software, and three SotA Transformer-based NMT models developed in an academic context including Marian~\cite{tiedemann-thottingal-2020-opus,junczys-dowmunt-etal-2018-marian-cost}, mBART50~\cite{liu-etal-2020-multilingual-denoising,tang-etal-2021-multilingual}, and M2M100~\cite{JMLR:v22:20-1307}. We leverage them to generate the $1$-best translations for \textsc{Translate Train} and \textsc{Translate Test}, and we also compare the BLEU scores of the MT systems against the TTI performance. Note that training a \textsc{Translate Train} TTI model from scratch for each of the MT systems also takes $75$ hours; our \textsc{Translate Train} experiments thus do not extend to other datasets beyond COCO-CN due to the high computational cost. 

Given the considerations above along with preliminary evaluations on COCO-CN which showed that Marian outperforms mBART50 and M2M100, for the other datasets we focus on comparing the Marian-based \textsc{Translate Test} with \textsc{Zero-Shot Transfer}.

\rparagraph{RQ2 Experiments} RQ2 further studies the effectiveness of the proposed \method module; see \S\ref{s:methodology} and Figure~\ref{fig:arch}. We select Marian as the NMT backbone\footnote{Amazon Translate's API can only return a single $1$-best translation; it thus cannot be used for \method experiments.} and sample $m$ \textsc{en} translations per each input sentence in the input language $\mathbb{L}$.\footnote{We adopt the common \href{https://huggingface.co/docs/transformers/v4.20.1/en/main_classes/text_generation\#transformers.generation_utils.GenerationMixin.beam_sample}{`beam search multinomial sampling'} for sampling \textsc{en} translations.} To compare with \method (with the frozen mLAFITE generator), we also propose and experiment with several insightful and simple baselines (without the use of \method) in addition to the RQ1 baselines: \textbf{1)} we try standard mean-pooling as a simple ensembling baseline directly on mLAFITE; \textbf{2)} we fine-tune $G$ using the original non-\textsc{en} captions;\footnote{$G$ is tuned adversarially following the original training setup of {(m)LAFITE}.} \textbf{3)} we fine-tune $G$ using mean-pooled text features. Finally, we also investigate variants which combine \mbox{\method} with the tunable generator $G$ to check if further gains can be achieved.\footnote{We first fine-tune $G$ and then train \method with the fine-tuned $G$ but still use the discriminator of our pretrained mLAFITE to alleviate its overfitting~\cite{NEURIPS2020_8d30aa96}. We also tried \textbf{1)} training \method first and then fine-tuning $G$ and \textbf{2)} training \method together with $G$, but they both
derive suboptimal results in our preliminary investigation.}


Training for RQ2 experiments is conducted on $8$$\times$V100 GPUs with a batch size per GPU of $16$ for about $7$ hours (i.e., a total of $2$ million data points sampled from the respective training sets). We use Adam optimiser~\cite{Kingma2014AdamAM} with a learning rate of $5$e-$4$ and betas of $(0,0.99)$. For the generator-tuning baselines, their contrastive loss setups completely follow the original LAFITE~\cite{Zhou_2022_CVPR}. In our \method experiments, $\lambda_{1}$$=$$4$ and $\lambda_{2}$$=$$2$. Other hyper-parameters are as follows: the NMT beam size is $12$, NMT temperature is $2.0$, images are scaled to resolution $256\times256$, $m$$=$$12$, $d$$=$$512$, $d_{hid}$=$256$, and $\tau$$=$$0.5$. In addition, we fuse $10\%$ and $1\%$ standard Gaussian noise into $\mathbf{h}^{0}$ and $\mathbf{h}^{i} (1\leq i\leq m$) respectively as a data augmentation `trick'. The hyper-parameters are tuned on our dev split of COCO-CN with details in Appendix~\ref{appendix:reproducibility}. The same set of hyper-parameters is also adopted for the other two datasets. 


\rparagraph{Side Experiments} Besides the main RQ1 and RQ2 experiments, we also conduct a series of side analyses focused on \method. They span \textbf{1)} the impact of the number of \textsc{en} translations $m$, \textbf{2)} the impact of the interpolation hyperparameter $\alpha$, and \textbf{3)} robustness tests. We also conduct \textbf{4)} ablation studies to validate the effectiveness of different components, and \textbf{5)} present generated images and \method attention scores.

\rparagraph{Evaluation Metric} Following \newcite{Zhou_2022_CVPR} and \newcite{pmlr-v139-ramesh21a}, we report the Fr\'echet Inception Distance (FID)~\cite{NIPS2017_8a1d6947} computed with $30,000$ synthesised images generated using randomly sampled test set texts against test set ground-truth images, which is the most authoritative machine evaluation metric for TTI so far.\footnote{Inception
Score (IS)~\cite{NIPS2016_8a3363ab} is another common evaluation metric for TTI, which is, to some extent, superseded by FID~\cite{NIPS2017_8a1d6947,BORJI2022103329}. Moreover, IS gives misleading results when applied to datasets other than ImageNet~\cite{Barratt2018ANO}, and is especially non-fitting for unannotated LAION-5B images.}


\section{Results and Discussion}
\label{s:results}
The main results are structured around the two central RQs from \S\ref{s:introduction}, discussed in \S\ref{s:RQ1results} and \S\ref{s:RQ2results}.

\subsection{RQ1: Results and Analyses}
\label{s:RQ1results}
\sparagraph{Comparison of Three Baselines} The results of \textsc{Translate Train}, \textsc{Translate Test}, and \textsc{Zero-Shot Transfer} on COCO-CN are summarised in Table~\ref{table:RQ1Results1}. While all three methods use mCLIP, \textsc{Translate Test} and \textsc{Zero-Shot Transfer} are based on a pretrained \textsc{en} mLAFITE and do not require any further tuning. \textsc{Translate Train} achieves the best FID scores; however, it requires training from scratch with translated $\mathbb{L}$ captions (see \S\ref{s:methodology_mlafite} and \S\ref{s:experiments}). Since MS-COCO provides ground-truth human-written \textsc{en} captions for COCO-CN images, and Multi30K Task2 also provides \textsc{en} human descriptions, we directly feed the \textsc{en} captions to mLAFITE and report the FID scores as an upper bound (see the first row of each of Tables~\ref{table:RQ1Results1} and~\ref{table:RQ1Results2}).\footnote{For fair comparisons, we keep only one ground-truth \textsc{en} caption for each image.} 


The scores in Tables~\ref{table:RQ1Results1} and~\ref{table:RQ1Results2} show that \textsc{Zero-Shot Transfer} outperforms \textsc{Translate Test}, demonstrating the strong capability of the multilingual mCLIP text encoder. \textsc{Translate Test} compares unfavourably to other methods, revealing the gap between \textsc{en} translations and the ground-truth \textsc{en} human descriptions (e.g., translation errors, `translationese' bias). We further extend the comparison to five more languages from the IGLUE dataset, and the results from Table~\ref{table:RQ1Results3} in Appendix~\ref{appendix:IGLUE} corroborate the finding that \textsc{Zero-Shot Transfer} generally outperforms \textsc{Translate Test}.  

\begin{table}[!t]
\centering
\def\arraystretch{0.83}
\resizebox{0.99\linewidth}{!}{%
\begin{tabular}{llll}
\toprule 
\rowcolor{Gray}
\multicolumn{1}{c}{\bf Method}  &\multicolumn{1}{c}{\bf MT Model}  &\multicolumn{1}{c}{\bf BLEU $\uparrow$} &\multicolumn{1}{c}{\bf FID $\downarrow$}\\
\midrule
\multirow{1}{4.5cm}{\centering Ground-Truth \textsc{en} Captions}&\multicolumn{1}{c}{-}&\multicolumn{1}{c}{-} &\multicolumn{1}{c}{14.35}\\
\midrule
\multirow{4}{4.5cm}{\centering \textsc{Translate Train\\} (\textsc{en}$\to$\textsc{zh})}&\multicolumn{1}{c}{mBART50
} &\multicolumn{1}{c}{32.77} &\multicolumn{1}{c}{14.98}\\
&\multicolumn{1}{c}{Marian}  &\multicolumn{1}{c}{32.5 }&\multicolumn{1}{c}{\bf 14.64}\\
&\multicolumn{1}{c}{M2M100}  &\multicolumn{1}{c}{33.73 }&\multicolumn{1}{c}{15.28}\\
&\multicolumn{1}{c}{Amazon Translate}  &\multicolumn{1}{c}{\bf 42.23 }&\multicolumn{1}{c}{14.87}\\
\midrule
\multirow{4}{4.5cm}{\centering \textsc{Translate Test\\} (\textsc{zh}$\to$\textsc{en})}&\multicolumn{1}{c}{mBART50
} &\multicolumn{1}{c}{26.32} &\multicolumn{1}{c}{16.38}\\
&\multicolumn{1}{c}{Marian}  &\multicolumn{1}{c}{25.11}&\multicolumn{1}{c}{15.9}\\
&\multicolumn{1}{c}{M2M100}  &\multicolumn{1}{c}{22.65}&\multicolumn{1}{c}{17.26}\\
&\multicolumn{1}{c}{Amazon Translate}  &\multicolumn{1}{c}{\bf 30.95}&\multicolumn{1}{c}{\bf 15.64}\\
\midrule
\multirow{1}{4.5cm}{\centering\textsc{Zero-Shot Transfer}}&\multicolumn{1}{c}{-}&\multicolumn{1}{c}{-} &\multicolumn{1}{c}{15.57}\\

\bottomrule
\end{tabular}
}
\caption{Results on COCO-CN (\textsc{zh}). `-': the method does not rely on MT. \textbf{FID}$\downarrow$: lower is better.}
\label{table:RQ1Results1}
\end{table}

\begin{table}[!t]
\def\arraystretch{0.83}
\centering
\resizebox{0.99\linewidth}{!}{%
\begin{tabular}{llll}
\toprule 
\rowcolor{Gray}
\multicolumn{1}{c}{\bf Method}  &\multicolumn{1}{c}{\bf \textsc{zh}: FID $\downarrow$}  &\multicolumn{1}{c}{\bf \textsc{de}: FID $\downarrow$} &\multicolumn{1}{c}{\bf\textsc{fi}: FID $\downarrow$}\\
\midrule
\multirow{1}{6cm}{\centering Ground-Truth \textsc{en} Captions}&\multicolumn{1}{c}{14.35}&\multicolumn{1}{c}{16.68} &\multicolumn{1}{c}{-}\\
\midrule
\multirow{1}{6cm}{\centering \textsc{Translate Test} (Marian)}&\multicolumn{1}{c}{15.9}&\multicolumn{1}{c}{17.31} &\multicolumn{1}{c}{27.23}\\
\multirow{1}{6cm}{\centering \textsc{Translate Test} (Amazon Translate)}&\multicolumn{1}{c}{15.64}&\multicolumn{1}{c}{17.03} &\multicolumn{1}{c}{26.67}\\

\multirow{1}{6cm}{\centering\textsc{Zero-Shot Transfer}}&\multicolumn{1}{c}{\bf 15.57}&\multicolumn{1}{c}{\bf 16.98} &\multicolumn{1}{c}{\bf 25.78}\\

\bottomrule
\end{tabular}
}
\caption{\textsc{Translate Test} vs. \textsc{Zero-Shot Transfer} on COCO-CN (\textsc{zh}), Multi30K Task2 (\textsc{de}), and LAION-5B 
 (\textsc{fi}). `-': LAION-5B (\textsc{fi}) data do not provide ground-truth \textsc{en} captions. \textbf{FID}$\downarrow$: lower is better.} 
\label{table:RQ1Results2}
\end{table}

\rparagraph{Comparison of MT Systems}
We compare the performance of the four MT systems on COCO-CN and also report their BLEU scores on the additional $5K$ sentence pairs. Table~\ref{table:RQ1Results1}, as expected, reveals that the commercial Amazon Translate system offers much stronger MT performance than the three academic NMT systems in terms of BLEU. Concerning mTTI, Amazon Translate is the best system with the \textsc{Translate Test} approach category and ranks second with \textsc{Translate Train}. Interestingly, there are some salient discrepancies between BLEU-based versus TTI-based system rankings. For example, Marian ranks second in \textsc{Translate Test} and is the best system with \textsc{Translate Train}, although its MT performance underperforms both Amazon Translate and mBART50. We speculate that this might be due to the pretraining specifics of mCLIP, where \textit{Marian-generated pseudo-parallel sentence pairs} were used~\cite{carlsson-etal-2022-cross}. 

In \textsc{Translate Test}, M2M100 obtains the lowest \textsc{zh}$\to$\textsc{en} BLEU score and also achieves the worst TTI performance. However, mBART50 and M2M100 have close \textsc{en}$\to$\textsc{zh} BLEU scores in \textsc{Translate Train}, and a small edge in BLEU cannot guarantee a better TTI performance. We additionally compare Marian and Amazon Translate for \textsc{Translate Test} in Tables~\ref{table:RQ1Results2} and~\ref{table:RQ1Results3} (Appendix~\ref{appendix:IGLUE}) on other languages and datasets, which further validate the core findings.



\subsection{RQ2: Results and Analyses}
\label{s:RQ2results}
\sparagraph{Effectiveness of \method}
The main results are summarised in Table~\ref{table:RQ2Results}. 
%
%
%
For all methods except `Ground-Truth \textsc{en} Captions', the language gap (with \textsc{en} captions for mLAFITE pretraining) always exists since the text input is in language $\mathbb{L}$. When there is no image domain gap (i.e., for COCO-CN), \method without tuning $G$ achieves the best score, surpassing also the \textsc{Translate Train} baseline (cf.\;Table~\ref{table:RQ1Results1}), and the absolute score also mitigates the gap to the upper-bound set by `Ground-Truth \textsc{en} Captions'. With image domain gap present (i.e., \textsc{de} and \textsc{fi}), training \method (with frozen $G$) still shows a small edge over fine-tuning $G$ (without \method) for \textsc{de}; however, for the noisier LAION-5B data, fine-tuning $G$ is more useful. However, for both \textsc{de} and \textsc{fi}, the best results are always achieved when \method is leveraged, validating its usefulness combined with parameter efficiency. For example, \method with $G$ frozen consistently outperforms \textsc{Zero-Shot Transfer} while introducing only $0.1$\% extra parameters. Our robustness tests repeating \method (Frozen $G$) experiments on COCO-CN with different random seeds further corroborate these findings (the deviation of FID is $0.04$), with a short summary in Appendix~\ref{app:robustness}.



\begin{table}[!t]
\centering
\resizebox{1.0\linewidth}{!}{%
\begin{tabular}{llll}
\toprule 
\rowcolor{Gray}
\multicolumn{1}{c}{\bf Method}  &\multicolumn{1}{c}{\bf \textsc{zh}: FID $\downarrow$}  &\multicolumn{1}{c}{\bf \textsc{de}: FID $\downarrow$} &\multicolumn{1}{c}{\bf\textsc{fi}: FID $\downarrow$}\\
\midrule
\multirow{1}{6cm}{\centering Ground-Truth \textsc{en} Captions}&\multicolumn{1}{c}{14.35}&\multicolumn{1}{c}{16.68} &\multicolumn{1}{c}{-}\\

\multirow{1}{6cm}{\centering\textsc{Zero-Shot Transfer}}&\multicolumn{1}{c}{15.57}&\multicolumn{1}{c}{16.98} &\multicolumn{1}{c}{25.78}\\

\multirow{1}{6cm}{\centering Mean Pooling}&\multicolumn{1}{c}{ 16.47}&\multicolumn{1}{c}{17.7} &\multicolumn{1}{c}{27.67}\\
\midrule
\multirow{1}{6cm}{\centering Fine-Tune $G$ ($\mathbb{L}$ Text)}&\multicolumn{1}{c}{15.23}&\multicolumn{1}{c}{16.28} &\multicolumn{1}{c}{17.69}\\

\multirow{1}{6cm}{\centering Fine-Tune $G$ (Mean Pooling)}&\multicolumn{1}{c}{15.27}&\multicolumn{1}{c}{16.68} &\multicolumn{1}{c}{18.17}\\
\midrule

\multirow{1}{6cm}{\centering \method (Frozen $G$)}&\multicolumn{1}{c}{\bf \; \; \; \; 14.52 \small$\downarrow^{6.7\%}$}&\multicolumn{1}{c}{16.26} &\multicolumn{1}{c}{21.9}\\

\multirow{1}{6cm}{\centering \method + Fine-Tune $G$ ($\mathbb{L}$ Text) }&\multicolumn{1}{c}{15.14}&\multicolumn{1}{c}{\bf \; \; \; \; 16.12 \small$\downarrow^{5.1\%}$} &\multicolumn{1}{c}{\bf \; \; \; \; \; 17.38 \small$\downarrow^{35.6\%}$}\\

\multirow{1}{6cm}{\centering \method + Fine-Tune $G$ (Mean Pooling)}&\multicolumn{1}{c}{14.93}&\multicolumn{1}{c}{16.23} &\multicolumn{1}{c}{17.41}\\

\bottomrule
\end{tabular}
}
\caption{Main results for RQ2. The models in the first three rows do not require any additional fine-tuning. For the best-performing models in \textbf{bold} numbers, we also present the relative improvement in percentage when comparing with \textsc{Zero-Shot Transfer}. \textbf{FID}$\downarrow$: lower is better.} 
\label{table:RQ2Results}
\end{table}
\rparagraph{Variants of \method}
We further investigate the impact of crucial design choices and hyper-parameters in \method such as $m$, $\alpha$, and $\bm{V}$ (see Eq.~\eqref{formula:Attn_QKV3}) respectively on the final TTI performance. The results of different variants are provided in Table~\ref{table:Variants}. They indicate that increasing the number of translations $m$ seems to be conducive to downstream TTI performance. In addition, when $\bm{V}=\bm{K}$, the FID score worsens, demonstrating the usefulness of the $\bm{V}$ variant as formulated by Eq.~\eqref{formula:Attn_QKV3}. Finally, the TTI performance deteriorates when $\alpha>0.2$, showing that $\mathbf{h}^{0}$ should still be the main component of $\tilde{\mathbf{h}}$, and \method provides auxiliary information (i.e., a translation-based enhancement).


\begin{table}[!t]
\centering
\def\arraystretch{0.93}
\resizebox{0.88\linewidth}{!}{%
\begin{tabular}{llll}
\toprule 
\rowcolor{Gray}
\multicolumn{1}{c}{\bf Model (Variant)}  &\multicolumn{1}{c}{\bf FID $\downarrow$}  &\multicolumn{1}{c}{\bf \bf Model (Variant)} &\multicolumn{1}{c}{\bf FID $\downarrow$}\\
\cmidrule(lr){1-2} \cmidrule(lr){3-4}

\multicolumn{1}{c}{Default}&\multicolumn{1}{c}{\bf 14.52 }&\multicolumn{1}{c}{Variant $4$: $\bm{V}=\bm{K}$} &\multicolumn{1}{c}{$14.73$}\\

\cmidrule(lr){1-2} \cmidrule(lr){3-4}
\multicolumn{1}{c}{\centering Variant $1$: $m=1$}&\multicolumn{1}{c}{ 14.9}&\multicolumn{1}{c}{Variant $5$: $\alpha=0.1$} &\multicolumn{1}{c}{15.07}\\

\multicolumn{1}{c}{\centering Variant $2$: $m=4$}&\multicolumn{1}{c}{ 14.65}&\multicolumn{1}{c}{Default: $\alpha=0.2$} &\multicolumn{1}{c}{\bf 14.52}\\

\multicolumn{1}{c}{\centering Variant $3$: $m=8$}&\multicolumn{1}{c}{14.68}&\multicolumn{1}{c}{Variant $6$: $\alpha=0.3$} &\multicolumn{1}{c}{14.81}\\

\multicolumn{1}{c}{\centering Default: $m=12$}&\multicolumn{1}{c}{\bf 14.52}&\multicolumn{1}{c}{Variant $7$: $\alpha=0.5$} &\multicolumn{1}{c}{17.7}\\

\bottomrule
\end{tabular}
}
\caption{Model variants of \method (Frozen $G$). FID scores on COCO-CN.} 
\label{table:Variants}
\end{table}

\rparagraph{Ablation Study} We now study the usefulness of two used contrastive losses: \textbf{1)} our proposed $\mathcal{L}_{CL}$ and \textbf{2)} $\mathcal{L}_{CL}^{D}$ inherited from LAFITE. The results in Table~\ref{table:ablation} show that removing $\mathcal{L}_{CL}$ causes a noticeable performance drop (increased FID). However, removing $\mathcal{L}_{CL}^{D}$ has only a minor impact on the FID score. When removing both CL losses, the adversarial losses alone produce an FID score of $14.82$. We also additionally try the CL loss setup of the original LAFITE and find that the setup is detrimental to the training of \method, producing a worse FID score than using the adversarial losses alone.

\begin{table}[!t]
\centering
\def\arraystretch{0.98}
\resizebox{0.82\linewidth}{!}{%
\begin{tabular}{ll}
\toprule 
\rowcolor{Gray}
\multicolumn{1}{c}{\bf Model (Variant)}  &\multicolumn{1}{c}{\bf FID $\downarrow$}\\
\midrule

\multicolumn{1}{c}{Default:  with $\mathcal{L}_{CL}$ and $\mathcal{L}_{CL}^{D}$}&\multicolumn{1}{c}{\bf 14.52}\\

\multicolumn{1}{c}{Remove $\mathcal{L}_{CL}$}&\multicolumn{1}{c}{14.74}\\

\multicolumn{1}{c}{Remove $\mathcal{L}_{CL}^{D}$}&\multicolumn{1}{c}{14.56}\\

\multicolumn{1}{c}{Remove both $\mathcal{L}_{CL}$ and  $\mathcal{L}_{CL}^{D}$}&\multicolumn{1}{c}{14.82}\\

\midrule 
\multicolumn{1}{c}{Setup of (m)LAFITE: with $\mathcal{L}_{CL}^{G}$ and $\mathcal{L}_{CL}^{D}$}&\multicolumn{1}{c}{15.03}\\

\bottomrule
\end{tabular}
}
\caption{Ablation study on CL losses. Model variant: \method (Frozen $G$). FID scores on COCO-CN.} 
\label{table:ablation}
\end{table}

\rparagraph{TTI Examples and Attention Scores}
Finally, we refer the reader to Appendix~\ref{appendix:examples} where we present images synthesised with \textsc{Translate Test}, \textsc{Zero-Shot Transfer}, and our \method models and where we also show the \method attention scores. The differences between images are subtle and we were unable to find a clear pattern that links high attention scores with particular translations.



\section{Conclusion}
\label{s:Conclution}
This work is one of the first investigations of multilingual and cross-lingual text-to-image generation (TTI), with a particular focus on investigating the use of machine translation (MT) for the task. We systematically compared standard cross-lingual transfer approaches \textsc{Translate Train}, \textsc{Translate Test} and \textsc{Zero-Shot Transfer} in the context of TTI and also studied the differences over MT systems. We
then proposed a novel Ensemble Adapter (\method) method that leverages multiple translations to further improve the TTI performance, with strong and consistent gains reported across a series of standard TTI benchmarks in different languages. 


\section*{Limitations}
\label{s:limitations}
First, we again emphasise that the lack of high-quality non-English image-caption pairs is a primary obstacle to wider-scale multilingual and cross-lingual TTI investigations. We hope that researchers in the future can construct and release more high-quality vision-language data for different languages, especially for low-resource ones.

Second, our work uses $512$-dim `XLM-R Large Vit-B/32' mCLIP\footnote{\href{https://github.com/FreddeFrallan/Multilingual-CLIP}{https://github.com/FreddeFrallan/Multilingual-CLIP}} and is based on the StyleGAN2 framework~\cite{Karras_2020_CVPR}. Since the main focus of our work is to realise multilingual and cross-lingual TTI and enable fair comparisons across different models and approaches, we compare all proposed and baseline methods with the same mCLIP text encoder and the GAN framework.
However, for readers and potential users interested in `chasing' stronger absolute FID scores, we speculate that the larger $640$-dim `XLM-R Large Vit-B/16+' mCLIP text encoder and the more recent StyleGAN3~\cite{NEURIPS2021_076ccd93} can be helpful.   

Third, we notice that in addition to LAFITE, several state-of-the-art large diffusion models such as those from \newcite{Saharia2022PhotorealisticTD} and~\newcite{Rombach_2022_CVPR} also use CLIP to condition image generation on text input. This means that we could be able to derive multilingual diffusion models for mTTI also by replacing CLIP with mCLIP and enhance the mTTI performance with our proposed \method (of course, we would need to redesign our loss functions). However, due to limited computational resources, we leave it to future work. 

Fourth, the \method boosts cross-lingual transfer for TTI by combining the knowledge from multiple translations, which can mitigate potential translation errors. Our work does not demonstrate if \method is applicable and adaptable to downstream cross-lingual tasks besides TTI. It is because \textbf{1)} downstream tasks other than TTI are out of the scope of this work and \textbf{2)} adapting \method to different tasks will require redesign of model structures and losses catering to the characteristics of each downstream task, making us believe it is not proper to expand the topic and include everything in a single piece of work. Therefore, we also leave this to future work.

\section*{Ethics Statement}
\label{s:ethics}
The datasets involved in our experiments are publicly available and widely used, and it is quite common to train text-to-image generation models on publicly available data. To the best of our knowledge, the ethical risk is minimal. For privacy concerns, we do not present images with human faces and captions with real human names in the paper, and we will not release material that may contain any sensitive information.  

\section*{Acknowledgements}
\label{s:acknowledgements}
We would like to thank \textbf{1)} all members of the Amazon Alexa Translations Science Team for helpful discussions and valuable comments during the weekly group meetings, \textbf{2)} Yufan Zhou, the author of LAFITE, who kindly responded to our questions concerning LAFITE's technical details on Github, and \textbf{3)} the anonymous reviewers for their feedback.

Ivan Vuli\'{c} is supported by a personal Royal Society University Research Fellowship \textit{`Inclusive and Sustainable Language Technology for a Truly Multilingual World'} (no 221137; 2022--).

\bibliography{custom}
\bibliographystyle{acl_natbib}

\clearpage
\appendix
\newpage
\section{Data Statistics and Languages}
\label{appendix:languages}
In Table~\ref{table:languages_details}, we summarise the data statistics and languages covered in our experiments.

\begin{table*}[ht!]
\begin{center}
\resizebox{0.99\linewidth}{!}{%
\begin{tabular}{lllllllllll}
\toprule 
\rowcolor{Gray}
\multicolumn{1}{c}{\bf Language}  &\multicolumn{1}{c}{\bf Family} &\multicolumn{1}{c}{\bf Code}   &\multicolumn{1}{c}{\bf Dataset} &\multicolumn{1}{c}{\bf Training Set: $\#$ of Images} &\multicolumn{1}{c}{\bf Dev Set: $\#$ of Images} &\multicolumn{1}{c}{\bf Test Set: $\#$ of Images} &\multicolumn{1}{c}{\bf Min Seq Len}  &\multicolumn{1}{c}{\bf Max Seq Len}  &\multicolumn{1}{c}{\bf Avg. Seq Len} &\multicolumn{1}{c}{\bf Image Domain Overlap}\\

\midrule
\multirow{1}{*}{English} &\multicolumn{1}{c}{Germanic} &\multicolumn{1}{c}{\textsc{en}} &\multicolumn{1}{c}{MS-COCO} &\multicolumn{1}{c}{74,505} &\multicolumn{1}{c}{8,278} &\multicolumn{1}{c}{40,504} &\multicolumn{1}{c}{5} &\multicolumn{1}{c}{50} &\multicolumn{1}{c}{10.5} &\multicolumn{1}{c}{-}\\

\midrule
\multirow{1}{*}{Chinese} &\multicolumn{1}{c}{Sino-Tibetan} &\multicolumn{1}{c}{\textsc{zh}} &\multicolumn{1}{c}{COCO-CN} &\multicolumn{1}{c}{10,875} &\multicolumn{1}{c}{2,718} &\multicolumn{1}{c}{6,748} &\multicolumn{1}{c}{5} &\multicolumn{1}{c}{63} &\multicolumn{1}{c}{17.3} &\multicolumn{1}{c}{\checkmark}\\

\midrule
\multirow{1}{*}{German} &\multicolumn{1}{c}{Germanic} &\multicolumn{1}{c}{\textsc{de}} &\multicolumn{1}{c}{Multi30K Task2} &\multicolumn{1}{c}{10,000} &\multicolumn{1}{c}{2,000} &\multicolumn{1}{c}{19,014} &\multicolumn{1}{c}{1} &\multicolumn{1}{c}{34} &\multicolumn{1}{c}{8.2} &\multicolumn{1}{c}{\text{\sffamily x}}\\

\midrule
\multirow{1}{*}{Finnish} &\multicolumn{1}{c}{Uralic} &\multicolumn{1}{c}{\textsc{fi}} &\multicolumn{1}{c}{LAION-5B} &\multicolumn{1}{c}{10,000} &\multicolumn{1}{c}{2,000} &\multicolumn{1}{c}{18,000} &\multicolumn{1}{c}{8} &\multicolumn{1}{c}{116} &\multicolumn{1}{c}{14.6} &\multicolumn{1}{c}{\text{\sffamily x}}\\

\midrule
\multirow{1}{*}{Spanish} &\multicolumn{1}{c}{Romance} &\multicolumn{1}{c}{\textsc{es}} &\multicolumn{1}{c}{IGLUE xFlickr$\&$CO} &\multicolumn{1}{c}{0} &\multicolumn{1}{c}{0} &\multicolumn{1}{c}{2,000} &\multicolumn{1}{c}{3} &\multicolumn{1}{c}{59} &\multicolumn{1}{c}{13.7} &\multicolumn{1}{c}{\notcheckmark}\\

\midrule
\multirow{1}{*}{Indonesian} &\multicolumn{1}{c}{Austronesian} &\multicolumn{1}{c}{\textsc{id}} &\multicolumn{1}{c}{IGLUE xFlickr$\&$CO} &\multicolumn{1}{c}{0} &\multicolumn{1}{c}{0} &\multicolumn{1}{c}{1,999} &\multicolumn{1}{c}{3} &\multicolumn{1}{c}{31} &\multicolumn{1}{c}{11.7} &\multicolumn{1}{c}{\notcheckmark}\\

\midrule
\multirow{1}{*}{Japanese} &\multicolumn{1}{c}{Japonic} &\multicolumn{1}{c}{\textsc{ja}} &\multicolumn{1}{c}{IGLUE xFlickr$\&$CO} &\multicolumn{1}{c}{0} &\multicolumn{1}{c}{0} &\multicolumn{1}{c}{2,000} &\multicolumn{1}{c}{5} &\multicolumn{1}{c}{175} &\multicolumn{1}{c}{33.8} &\multicolumn{1}{c}{\notcheckmark}\\

\midrule
\multirow{1}{*}{Russian} &\multicolumn{1}{c}{Slavic} &\multicolumn{1}{c}{\textsc{ru}} &\multicolumn{1}{c}{IGLUE xFlickr$\&$CO} &\multicolumn{1}{c}{0} &\multicolumn{1}{c}{0} &\multicolumn{1}{c}{2,000} &\multicolumn{1}{c}{1} &\multicolumn{1}{c}{45} &\multicolumn{1}{c}{11.3} &\multicolumn{1}{c}{\notcheckmark}\\

\midrule
\multirow{1}{*}{Turkish} &\multicolumn{1}{c}{Turkic} &\multicolumn{1}{c}{\textsc{tr}} &\multicolumn{1}{c}{IGLUE xFlickr$\&$CO} &\multicolumn{1}{c}{0} &\multicolumn{1}{c}{0} &\multicolumn{1}{c}{2,000} &\multicolumn{1}{c}{2} &\multicolumn{1}{c}{30} &\multicolumn{1}{c}{9.5} &\multicolumn{1}{c}{\notcheckmark}\\

\bottomrule
\end{tabular}
}

\caption{Data statistics categorised by languages. This table includes information such as language family, ISO 639-1 code, dataset name, train/dev/test split, and statistics on sequence length (number of words per caption). Note that MS-COCO \textsc{en} data is used for pretraining our mLAFITE only. We also show for each dataset if there is an image domain overlap with MS-COCO images used for mLAFITE pretraining. \checkmark: all images are from MS-COCO; \text{\sffamily x}: none of the images is from MS-COCO; \notcheckmark: half of the images are from MS-COCO. For IGLUE Indonesian data, we remove an empty caption and its associated image, so there are $1,999$ images left.}
\label{table:languages_details}
\end{center}
\end{table*}

\section{Additional Discussion on Data Sources}
\label{appendix:data}

Even without human-annotated image descriptions, there are two possible ways to derive captions for a target language $\mathbb{L}$.

First, we could translate \textsc{en} captions into $\mathbb{L}$ manually (still costly) or via machine translation. Our \textbf{\textsc{Translate Train}} baseline (see \S\ref{s:methodology}) derives training data via machine translation and trains an $\mathbb{L}$ TTI model from scratch. One main disadvantage of this approach is that it incurs huge training costs. While translations can be used as training data, we are conservative about using translated captions for TTI evaluation which can cause unexpected bias~\cite{elliott-etal-2016-multi30k,van-miltenburg-etal-2017-cross,pmlr-v162-bugliarello22a}. 

Second, it is possible to use cheaper but noisy Web-crawled visual-language data. For example, the recently released LAION-5B dataset~\cite{Schuhmann2022LAION5BAO} has 5 billion image-text pairs for $100$+ languages. There are previous examples that successfully trained SotA \textsc{en} TTI models with Web-crawled data, such as large VQVAE-based models and diffusion models. The models described in \newcite{pmlr-v139-ramesh21a},~\newcite{pmlr-v162-nichol22a} and~\newcite{Ramesh2022HierarchicalTI} are trained on \textsc{en} large-scale Web-crawled data, but are eventually also tested on the gold-standard MS-COCO validation set. In our work, in addition to two gold-standard datasets, we also try to build on our own a small-scale dataset for both training and evaluation by filtering relatively good-quality image-text pairs from a subset of the noisy LAION-5B data (details in \S\ref{s:datasets}). Training non-\textsc{en} TTI models from scratch with large-scale Web-crawled data such as LAION-5B is out of the scope of our work, and we focus on cross-lingual transfer learning setups with limited $\mathbb{L}$ data. As mentioned in \S\ref{s:introduction}, this is to a large extent due to concerns about huge computational costs for training TTI models. Moreover, there are circa $7,000$ languages worldwide~\cite{ethnologue}, and for low-resource languages not covered in LAION-5B's $100+$ languages, cross-lingual transfer learning approaches would still be the first choice. Furthermore, the number of \textsc{en} texts in LAION-5B is more than the total amount of texts from its 100+ non-\textsc{en} texts. Making full use of the huge amount of \textsc{en} image-text pairs via cross-lingual transfer learning might be beneficial for other languages. Therefore, we think that cross-lingual transfer learning in relatively low-resource setups for multilingual TTI is a critical and valuable research topic.

\section{The Detailed Training Process of \method}
\label{app:alg}
We summarise the training process of our \method method (see \S\ref{s:methodology}) in Algorithm~\ref{alg}.

\begin{algorithm}[!t]
\caption{{\footnotesize Supervised Training of Ensemble Adapter}}
{\footnotesize
\begin{algorithmic}[1]
\State \textbf{Input:} An image-text dataset $\{\mathbf{x}^{0}_{i},\mathcal{I}_{i}^{real}\}_{i=1}^{N}$
\State Derive $\bm{H}_{i}$ for each $\mathbf{x}^{0}_{i}$ with NMT and mCLIP
\State \textbf{while} \textit{not converge} \textbf{do:}
\State \quad Sample mini-batch $\{\bm{H}_{i},\mathcal{I}_{i}^{real}\}_{i=1}^{n}$;
\State \quad Sample random noise $\{\mathbf{z}_{i}\}_{i=1}^{n}\sim\mathcal{N}(\mathbf{0},\mathbf{I})$;
\State \quad \method forward pass $\tilde{\mathbf{h}}_{i}$$\gets$$\method(\bm{H}_{i})$;
\State \quad Synthesise fake image $\mathcal{I}^{fake}_{i}$$\gets$$G(\tilde{\mathbf{h}}_{i},\mathbf{z})$
\State \quad Feed $(\tilde{\mathbf{h}}_{i},\mathcal{I}^{real})$ and $(\tilde{\mathbf{h}}_{i},\mathcal{I}^{fake})$ to D respectively;
\State \quad Update \method with Eq.~\eqref{formula:Loss-EA};
\State \quad Update $D$ with Eq.~\eqref{formula:Loss-D};
\State \textbf{end while}
\end{algorithmic}
}%
\label{alg}
\end{algorithm}

\section{Deriving LAION-5B Dataset for Finnish}
\label{app:fi}
We download circa $5.1$ million image-caption pairs from the \textsc{fi} category of LAION-5B. Since the Web-crawled data are noisy, we apply several filtering steps: \textbf{1)} since our images will be scaled to resolution $256\times256$, to avoid distortion we keep only images with their width-height ratio between $0.5$ and $2$; \textbf{2)} we keep captions with a minimum length of $8$ words, which is also a requirement of MS-COCO~\cite{Chen2015MicrosoftCC} in its data annotation; \textbf{3)} we use the langdetect library\footnote{\href{https://pypi.org/project/langdetect/}{https://pypi.org/project/langdetect/}} to remove texts misclassified into the LAION-5B \textsc{fi} category and make sure the texts left are actually in Finnish; \textbf{4)} we keep captions with one ending period `.'.\footnote{In our initial trial, we found that, among the highest mCLIP-scored $30K$ pairs, most captions which do not end with `.' are noisy short ads.} After these steps, $239K$ pairs are left, and we calculate mCLIP scores (cosine similarities between mCLIP-extracted text and image features) for all the pairs and keep the $30K$ highest-ranking pairs as the final dataset. We randomly split the data into training, development and test portions with $10,000$, $2,000$, and $18,000$ pairs, respectively. 

We `sanity-check' $50$ randomly sampled instances from our filtered data and find that, in most cases, the text matches the image content. But there are a small number of exceptional cases where the text contains extra information beyond the image content itself  (e.g., event descriptions). Overall, the quality of our \textsc{fi} data still cannot match MS-COCO or Multi30K. Another interesting finding is that LAION-5B captions often use real and concrete names such as `Messi' and `the national stadium' to describe the image content, while MS-COCO and Multi30K tend to use general words such as `a man'/`a football player' and `a stadium'/`a building'.


\section{RQ1: Results on IGLUE}
\label{appendix:IGLUE}

Table~\ref{table:RQ1Results3} shows additional TTI results on five languages from IGLUE, comparing \textsc{Translate Test} (with Marian and Amazon Translate) and \textsc{Zero-Shot Transfer} baselines. 

\begin{table*}[ht!]
\begin{center}
\resizebox{0.92\linewidth}{!}{%
\begin{tabular}{lllllll}
\toprule 
\rowcolor{Gray}
\multicolumn{1}{c}{\bf Method}  &\multicolumn{1}{c}{\bf \textsc{es}: FID $\downarrow$}  &\multicolumn{1}{c}{\bf \textsc{id}: FID $\downarrow$} &\multicolumn{1}{c}{\bf\textsc{ja}: FID $\downarrow$} &\multicolumn{1}{c}{\bf\textsc{ru}: FID $\downarrow$} &\multicolumn{1}{c}{\bf\textsc{tr}: FID $\downarrow$} &\multicolumn{1}{c}{\bf Avg.: FID $\downarrow$}\\

\midrule
\multirow{1}{6cm}{\centering \textsc{Translate Test} (Marian)}&\multicolumn{1}{c}{\bf 30.04}&\multicolumn{1}{c}{31.09} &\multicolumn{1}{c}{33.12} &\multicolumn{1}{c}{31.11}&\multicolumn{1}{c}{30.74} &\multicolumn{1}{c}{31.22}\\

\multirow{1}{6cm}{\centering \textsc{Translate Test} (Amazon Translate)}&\multicolumn{1}{c}{30.08}&\multicolumn{1}{c}{31.27} &\multicolumn{1}{c}{\bf 31.61} &\multicolumn{1}{c}{30.83}&\multicolumn{1}{c}{30.31} &\multicolumn{1}{c}{30.82}\\

\multirow{1}{6cm}{\centering\textsc{Zero-Shot Transfer}}&\multicolumn{1}{c}{30.31}&\multicolumn{1}{c}{\bf 30.58} &\multicolumn{1}{c}{32.24} &\multicolumn{1}{c}{\bf 30.77}&\multicolumn{1}{c}{\bf 30.12} &\multicolumn{1}{c}{\bf 30.8}\\

\bottomrule
\end{tabular}
}

\caption{RQ1 results: \textsc{Translate Test} vs. \textsc{Zero-Shot Transfer} on five languages from IGLUE. \textbf{FID}$\downarrow$: lower is better.} 
\label{table:RQ1Results3}
\end{center}
\end{table*}

\section{Robustness of \method} 
\label{app:robustness}
We train the `\method (Frozen $G$)' model on COCO-CN $6$ more times ($7$ times in total) with different random seeds, and for each saved model we run TTI evaluation three times.\footnote{\textsc{de} and \textsc{fi} have larger test set sizes, so their evaluation results should be more reliable~\cite{Chong_2020_CVPR}.} Finally, we get $21$ FID results, with min $14.47$, max $14.62$, mean $14.55$, and standard deviation $0.04$. Even the worst score of $14.62$ outperforms all other baselines on COCO-CN.  


\section{Reproducibility Checklist}
\label{appendix:reproducibility}
\begin{itemize}

    \item \textbf{TTI Data}: the datasets used in our work are all publicly available including MS-COCO\footnote{\href{https://cocodataset.org/}{https://cocodataset.org}}, COCO-CN\footnote{\href{https://github.com/li-xirong/coco-cn}{https://github.com/li-xirong/coco-cn}}, Multi30K Task2\footnote{\href{https://github.com/multi30k/dataset}{https://github.com/multi30k/dataset}}, LAION-5B\footnote{\href{https://laion.ai/blog/laion-5b/}{https://laion.ai/blog/laion-5b}}, and IGLUE\footnote{\href{https://github.com/e-bug/iglue}{https://github.com/e-bug/iglue}}.
    \item \textbf{Parameter Counts}: the number of parameters is $655,873$ for our ensemble adapter network, $44,997,026$ for the generator network, $29,126,785$ for the discriminator network, $560,415,232$ for the mCLIP text encoder `M-CLIP/XLM-Roberta-Large-Vit-B-32'\footnote{\href{https://github.com/FreddeFrallan/Multilingual-CLIP}{https://github.com/FreddeFrallan/Multilingual-CLIP}}, and $87,849,216$ for the CLIP visual encoder `ViT-B/32'\footnote{\href{https://github.com/openai/CLIP}{https://github.com/openai/CLIP}}.
    \item \textbf{Computing Infrastructure}: we run our code on an Amazon EC2 P3.16xlarge Instance with $8$$\times$$16$GB Nvidia$^{\tiny\circledR}$ Tesla$^{\tiny\circledR}$ V100 GPUs, $64$$\times$$2.30$ GHz Intel$^{\tiny\circledR}$ Xeon$^{\tiny\circledR}$ E5-2686 v4 CPU cores, and $488$GB RAM. 
    \item \textbf{Software}: Python $3.7.0$, PyTorch $1.12.1$, and Transformers $4.21.0$.    
    \item \textbf{Hyperparameter Search}: our hyper-parameters are tuned on our dev split of COCO-CN. The same hyper-parameters are used for Multi30K and LAION-5B (we also conduct minimal tuning on their dev sets and find that the hyper-parameters tuned on COCO-CN are already (near-)optimal in our initial investigation). The learning rate is selected from $\{$$5$$e-$$5$$, 2.5$$e-$$4, 5$$e-$$4, 2.5$$e-$$3, 5$$e-$$3\}$, $\lambda_{1}$ and $\lambda_{2}$ which are weights for contrastive losses from $\{0.5, 1, 2, 4, 5, 10\}$, $\alpha$ the interpolation hyperparameter from $\{0.05,0.1,0.15,0.2,0.25,0.3,0.35,0.4,0.5\}$, and $d_{hid}$ from $\{32,64,128,256,512\}$. 
    \item \textbf{Runtime}: it takes $75$ hours to train an mLAFITE TTI model or a \textsc{Translate Train} model from scratch, $7$ hours to train an \method based on a pretrained mLAFITE, $7.5$ hours to fine-tune $G$ (without \method) based on a pretrained mLAFITE, and about 4 minutes to run FID evaluation for our TTI model with \method (data preprocessing, NMT, and mCLIP feature extraction excluded). All experiments and measurements were conducted on $8$$\times$$16$GB V100 GPUs. 
    \item \textbf{Other Technical Details}: we adopt the `exponential sharpening' for all contrastive losses as specified in LAFITE's supplementary material.\footnote{\href{https://openaccess.thecvf.com/content/CVPR2022/supplemental/Zhou_Towards_Language-Free_Training_CVPR_2022_supplemental.pdf}{https://openaccess.thecvf.com/content/CVPR2022/supplemental/Zhou\_Towards\_Language-Free\_Training\_CVPR\_2022\_supplemental.pdf}}
    \item \textbf{Carbon Footprint}: we estimate that \textbf{1)} training an mLAFITE TTI model or a \textsc{Translate Train} model from scratch can cause the emission of circa $56$$\sim$$67$-kg CO$_{2}$ equivalents; \textbf{2)} training an \method model would result in about $5$$\sim$$6$-kg CO$_{2}$ equivalents. These estimations are based on our computational infrastructure and a publicly available `machine learning emissions calculator'~\cite{luccioni2019quantifying}.\footnote{\href{https://mlco2.github.io/impact/\#compute}{https://mlco2.github.io/impact/\#compute}}
    
\end{itemize}

\section{TTI Examples and Attention Scores}
\label{appendix:examples}
\subsection{TTI Examples}

\begin{figure*}[t!]
\centering
\includegraphics[width=0.99\linewidth]{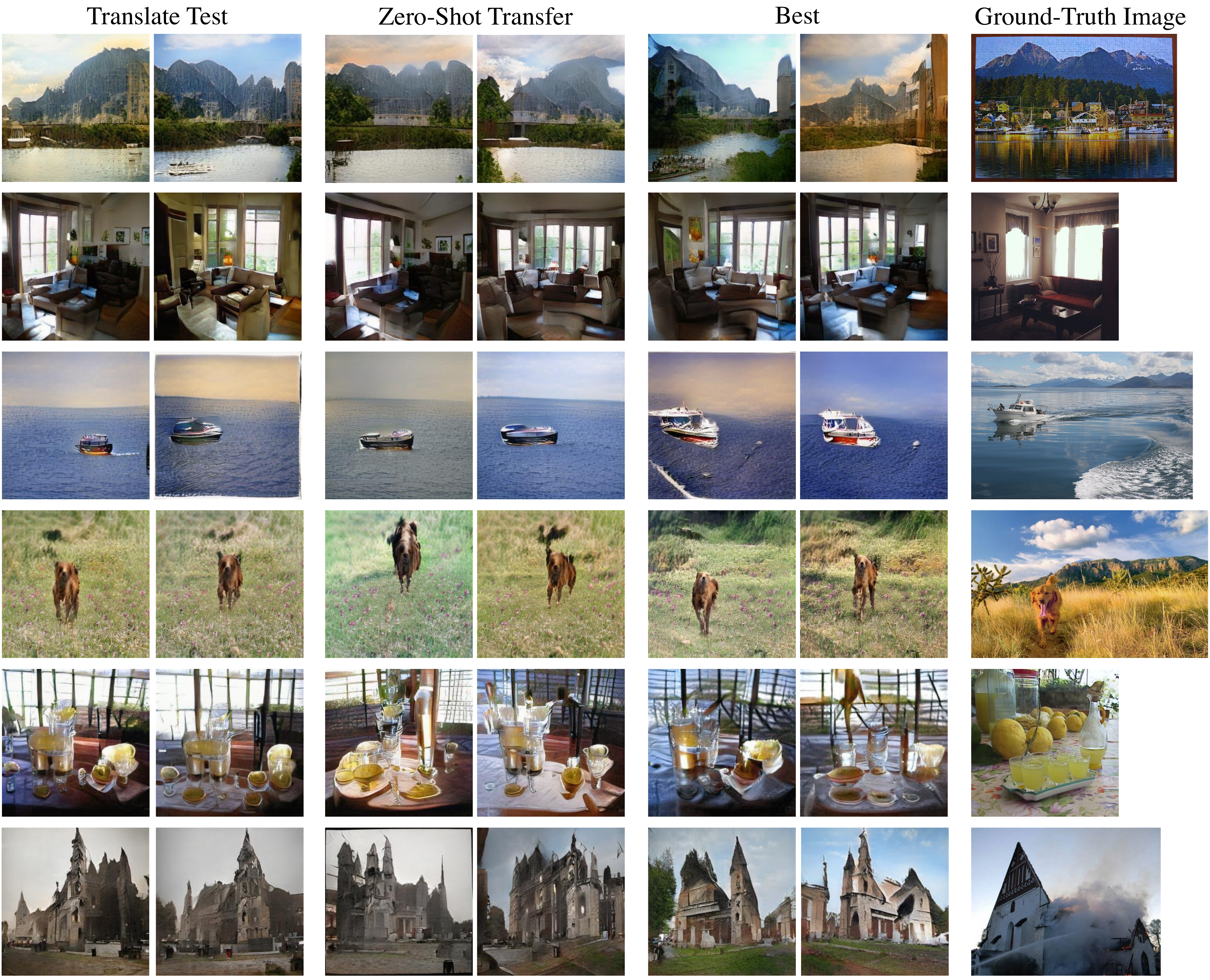}
\caption{TTI Examples generated with \textsc{Translate Test}, \textsc{Zero-Shot Transfer}, and our best model. COCO-CN (\textsc{zh}) Test Set: row $1$$-$$2$; Multi30K Task2 Test Set (\textsc{de}): row $3$$-$$4$; LAION-5B (\textsc{fi}) Test Set: row $5$$-$$6$. The resolution of the generated images is $256\times256$ pixels; ground-truth images are shown in their original sizes respectively.}
\label{fig:ttiexamples}
\end{figure*}

We compare images generated with \textsc{Translate Test}, \textsc{Zero-Shot Transfer}, and our best \method model in Figure~\ref{fig:ttiexamples}, where for each TTI method we present two images generated with different random noise inputs as introduced in \S\ref{s:methodology}. The `Best' model here refer to our \method model that achieve the best FID scores (\textbf{bold} numbers) in Table~\ref{table:RQ2Results} respectively for each language, i.e., `\method (Frozen $G$)' for \textsc{zh} and `\method + Fine-Tune $G$ ($\mathbb{L}$ Text)' for \textsc{de} and \textsc{fi}. The differences between images generated with different TTI methods are very subtle. 

\subsection{Attention Scores}
Table~\ref{table:attnscores} includes the original $\mathbb{L}$ input text, $\textsc{en}$ translations, and their associated \method attention scores (in descending order) corresponding to the images in Figure~\ref{fig:ttiexamples}. We did not identify any salient pattern concerning the type of \textsc{en} translations to which higher \method attention scores are attached.

\begin{table*}[ht!]
\begin{center}
\resizebox{0.97\linewidth}{!}{%
\begin{tabular}{lll}
\toprule 
\rowcolor{Gray}
\multicolumn{1}{c}{\bf Original $\mathbb{L}$ Input}  &\multicolumn{1}{c}{\bf \textsc{en} Translations}&\multicolumn{1}{c}{\bf \method Attention Scores}\\

\midrule
\multirow{12}{*}{\zh{房屋依水而建，远处群山环绕。}}
&\multicolumn{1}{c}{Houses are built with water, and they are surrounded by mountains at a distance.}&\multicolumn{1}{c}{8.75e-01}\\
&\multicolumn{1}{c}{Houses have been built waterly, surrounded by mountain ranges from one side to the other.}&\multicolumn{1}{c}{6.37e-02}\\
&\multicolumn{1}{c}{Houses are built on water, all around mountain mountains, as long as possible, and have access to water.}&\multicolumn{1}{c}{2.97e-02}\\
&\multicolumn{1}{c}{The house is constructed in the form of water, surrounded by mountains and long distances.}&\multicolumn{1}{c}{9.80e-03}\\
&\multicolumn{1}{c}{Houses are built with water and are located far beyond the range of hills around them.}&\multicolumn{1}{c}{9.17e-03}\\
&\multicolumn{1}{c}{The homes have been built on water and are surrounded by mountain areas from a distance.}&\multicolumn{1}{c}{5.64e-03}\\
&\multicolumn{1}{c}{Houses are built on water and surround it far from the mountains.}&\multicolumn{1}{c}{4.62e-03}\\
&\multicolumn{1}{c}{The houses are built according to water and spread around them from a great direction to a very deep range of mountains.}&\multicolumn{1}{c}{1.80e-03}\\
&\multicolumn{1}{c}{Houses are constructed around the mountain and built from a distant distance to an open point of view.}&\multicolumn{1}{c}{3.64e-04}\\
&\multicolumn{1}{c}{The houses are built by water and are encircled by mountains, as far as the hills are concerned.}&\multicolumn{1}{c}{2.27e-04}\\
&\multicolumn{1}{c}{The houses were built in the form of water. They were in a remote area around the mountains.}&\multicolumn{1}{c}{8.77e-05}\\
&\multicolumn{1}{c}{The houses were built watery and were driven from a very distant part of the forest and surrounded by mountains.}&\multicolumn{1}{c}{5.30e-08}\\
\midrule
\multirow{12}{*}{\zh{一个客厅，一个大窗户下面的沙发，桌子。}}
&\multicolumn{1}{c}{A living room, a couch under a huge window, a table.}&\multicolumn{1}{c}{3.08e-01}\\
&\multicolumn{1}{c}{A sitting room, a couch under a big window, a table.}&\multicolumn{1}{c}{2.24e-01}\\
&\multicolumn{1}{c}{A living room, a sofa under a big window, a table.}&\multicolumn{1}{c}{1.59e-01}\\
&\multicolumn{1}{c}{A living room, a couch under a big window, a table.}&\multicolumn{1}{c}{1.31e-01}\\
&\multicolumn{1}{c}{A living room, a couch under a big window, a table.}&\multicolumn{1}{c}{1.31e-01}\\
&\multicolumn{1}{c}{A living hall, a couch under that big window, a table.}&\multicolumn{1}{c}{2.12e-02}\\
&\multicolumn{1}{c}{I was in the living room, the sofa below the great window, the table.}&\multicolumn{1}{c}{1.27e-02}\\
&\multicolumn{1}{c}{In the living room, in the couch under a big window, in the table.}&\multicolumn{1}{c}{5.58e-03}\\
&\multicolumn{1}{c}{One living room. One large window under the couch. The table.}&\multicolumn{1}{c}{3.86e-03}\\
&\multicolumn{1}{c}{There was a living room, a couch under a large window, there was a table.}&\multicolumn{1}{c}{1.66e-03}\\
&\multicolumn{1}{c}{There's one room, and a big couch under the large window, and there's a table.}&\multicolumn{1}{c}{1.22e-03}\\
&\multicolumn{1}{c}{There was a guest room, there was a couch underneath a great window, there was a table.}&\multicolumn{1}{c}{7.53e-04}\\
\midrule
\multirow{12}{*}{Motorboot fährt auf ruhigem Gewässer}
&\multicolumn{1}{c}{Motor boat sails on calm waters}&\multicolumn{1}{c}{4.47e-01}\\
&\multicolumn{1}{c}{Motor boat sails on calm waters}&\multicolumn{1}{c}{4.47e-01}\\
&\multicolumn{1}{c}{Motor boat cruises on calm waters}&\multicolumn{1}{c}{6.43e-02}\\
&\multicolumn{1}{c}{Motorboat cruises on calm waters}&\multicolumn{1}{c}{2.14e-02}\\
&\multicolumn{1}{c}{Motor boat rides on calm waters}&\multicolumn{1}{c}{5.98e-03}\\
&\multicolumn{1}{c}{Motorboat travels on calm waters}&\multicolumn{1}{c}{5.46e-03}\\
&\multicolumn{1}{c}{Motorboat travels on calm waters}&\multicolumn{1}{c}{5.46e-03}\\
&\multicolumn{1}{c}{Motorboat drives on calm waters}&\multicolumn{1}{c}{1.60e-03}\\
&\multicolumn{1}{c}{Motorboat rides on calm waters}&\multicolumn{1}{c}{3.65e-04}\\
&\multicolumn{1}{c}{Motorboat rides on calm waters}&\multicolumn{1}{c}{3.65e-04}\\
&\multicolumn{1}{c}{Motorboat is sailing on calm waters}&\multicolumn{1}{c}{1.63e-04}\\
&\multicolumn{1}{c}{Motorboat is on the sea in order to keep its pace and to move towards the sea.}&\multicolumn{1}{c}{4.92e-05}\\
\midrule
\multirow{12}{*}{Einen braunen Hund der spazieren geht in der Wiese.}
&\multicolumn{1}{c}{I think he'd be able to walk in the meadow and we could have a brown dog to go for a walk.}&\multicolumn{1}{c}{5.72e-01}\\
&\multicolumn{1}{c}{He walks a brown dog in the meadows, who goes for a walk.}&\multicolumn{1}{c}{4.24e-01}\\
&\multicolumn{1}{c}{A brown dog that goes walking in the meadow.}&\multicolumn{1}{c}{1.88e-03}\\
&\multicolumn{1}{c}{A brown dog who goes for walks in the meadow.}&\multicolumn{1}{c}{1.16e-03}\\
&\multicolumn{1}{c}{A brown dog who goes for walks in the meadow.}&\multicolumn{1}{c}{1.16e-03}\\
&\multicolumn{1}{c}{A brown dog who goes for a walk in the meadow.}&\multicolumn{1}{c}{1.02e-04}\\
&\multicolumn{1}{c}{A brown dog going for a walk in the meadow.}&\multicolumn{1}{c}{1.10e-06}\\
&\multicolumn{1}{c}{A brown dog who walks in the meadow.}&\multicolumn{1}{c}{8.78e-09}\\
&\multicolumn{1}{c}{A brown dog taking a walk in the meadow.}&\multicolumn{1}{c}{4.56e-09}\\
&\multicolumn{1}{c}{A brown dog taking a walk in the meadow.}&\multicolumn{1}{c}{4.56e-09}\\
&\multicolumn{1}{c}{A brown dog walking in the meadow.}&\multicolumn{1}{c}{1.66e-09}\\
&\multicolumn{1}{c}{A brown dog walking in the meadow.}&\multicolumn{1}{c}{1.66e-09}\\
\midrule
\multirow{12}{*}{Terassin pöydällä kotitekoista limoncelloa lasipurkissa ja karahvissa sekä kaadettuna pieniin laseihin.}
&\multicolumn{1}{c}{Home made soda on the terrace in glass jar and karaaffles and poured into small glasses.}&\multicolumn{1}{c}{2.07e-01}\\
&\multicolumn{1}{c}{on the terrace table made homemade sodacello in a glass jar and in a girdle as well as poured into small glasses.}&\multicolumn{1}{c}{1.26e-01}\\
&\multicolumn{1}{c}{On the terrace table in a glass jar and a karahas of homemade limecello put down in small glasses.}&\multicolumn{1}{c}{8.66e-02}\\
&\multicolumn{1}{c}{On a terrace table of homemade lemonade in a glass jar and slab of gizzard and poured into small glasses.}&\multicolumn{1}{c}{8.64e-02}\\
&\multicolumn{1}{c}{The table on the terrace has homemade soda crystals in a glass jar and swath and is poured into small glasses.}&\multicolumn{1}{c}{8.30e-02}\\
&\multicolumn{1}{c}{On the terrace table housed lemonade in glass jars and swaths and poured down into small glasses.}&\multicolumn{1}{c}{8.24e-02}\\
&\multicolumn{1}{c}{On the terrace table of home made wine in glass jars and karaffes and poured into small glasses.}&\multicolumn{1}{c}{7.41e-02}\\
&\multicolumn{1}{c}{The terrace is equipped with homemade lemonade in the jar and perch and poured into small glasses.}&\multicolumn{1}{c}{5.89e-02}\\
&\multicolumn{1}{c}{Top of the terrace is homemade lemonade in a jar of glass and karaoke and poured into small glasses.}&\multicolumn{1}{c}{5.67e-02}\\
&\multicolumn{1}{c}{On the table of the terrace it's homemade limocello with glass pots and clovers and poured into small glasses.}&\multicolumn{1}{c}{5.50e-02}\\
&\multicolumn{1}{c}{On a table of terraces, homemade lemoncello is made in a glass jar and in a caraments and poured into small glasses.}&\multicolumn{1}{c}{4.91e-02}\\
&\multicolumn{1}{c}{on the table of terraces with homemade soda on a glass jar and karaffe and poured in small glasses.}&\multicolumn{1}{c}{3.56e-02}\\
\midrule
\multirow{12}{*}{Tuli tuhosi pahoin historiallisen kirkon vuoden 2006 toukokuussa.}
&\multicolumn{1}{c}{In May 2006 the historic church was badly destroyed by fire.}&\multicolumn{1}{c}{3.51e-01}\\
&\multicolumn{1}{c}{In May of 2006, the historical church was severely destroyed by fire.}&\multicolumn{1}{c}{2.39e-01}\\
&\multicolumn{1}{c}{It was, in May 2006, when the fire badly destroyed the historic church.}&\multicolumn{1}{c}{9.44e-02}\\
&\multicolumn{1}{c}{There was a great destruction of this historical church in May 2006.}&\multicolumn{1}{c}{6.56e-02}\\
&\multicolumn{1}{c}{The fire did a great deal of damage to an historic church in May 2006.}&\multicolumn{1}{c}{5.30e-02}\\
&\multicolumn{1}{c}{In May 2006 fire caused a very severe damage to the historic church.}&\multicolumn{1}{c}{3.58e-02}\\
&\multicolumn{1}{c}{The fire seriously destroyed the historical church in May 2006.}&\multicolumn{1}{c}{3.49e-02}\\
&\multicolumn{1}{c}{The fire was severely destroyed by the historical church in May 2006.}&\multicolumn{1}{c}{3.49e-02}\\
&\multicolumn{1}{c}{A fire severely destroyed the historical church in May 2006.}&\multicolumn{1}{c}{3.05e-02}\\
&\multicolumn{1}{c}{There's been massive damage to the historical Church in May 2006 when the fire took place.}&\multicolumn{1}{c}{3.01e-02}\\
&\multicolumn{1}{c}{Fire was devastatingly damaged by the historic church in May 2006.}&\multicolumn{1}{c}{2.18e-02}\\
&\multicolumn{1}{c}{Fire caused the serious destruction of the historic church in May 2006.}&\multicolumn{1}{c}{9.31e-03}\\

\bottomrule
\end{tabular}
}
\caption{\method attention scores.}
\label{table:attnscores}
\end{center}
\end{table*}

\subsection{Can \method Incorporate Manually Added Information from Translations?}
\begin{figure*}[t!]
\centering
\includegraphics[width=0.99\linewidth]{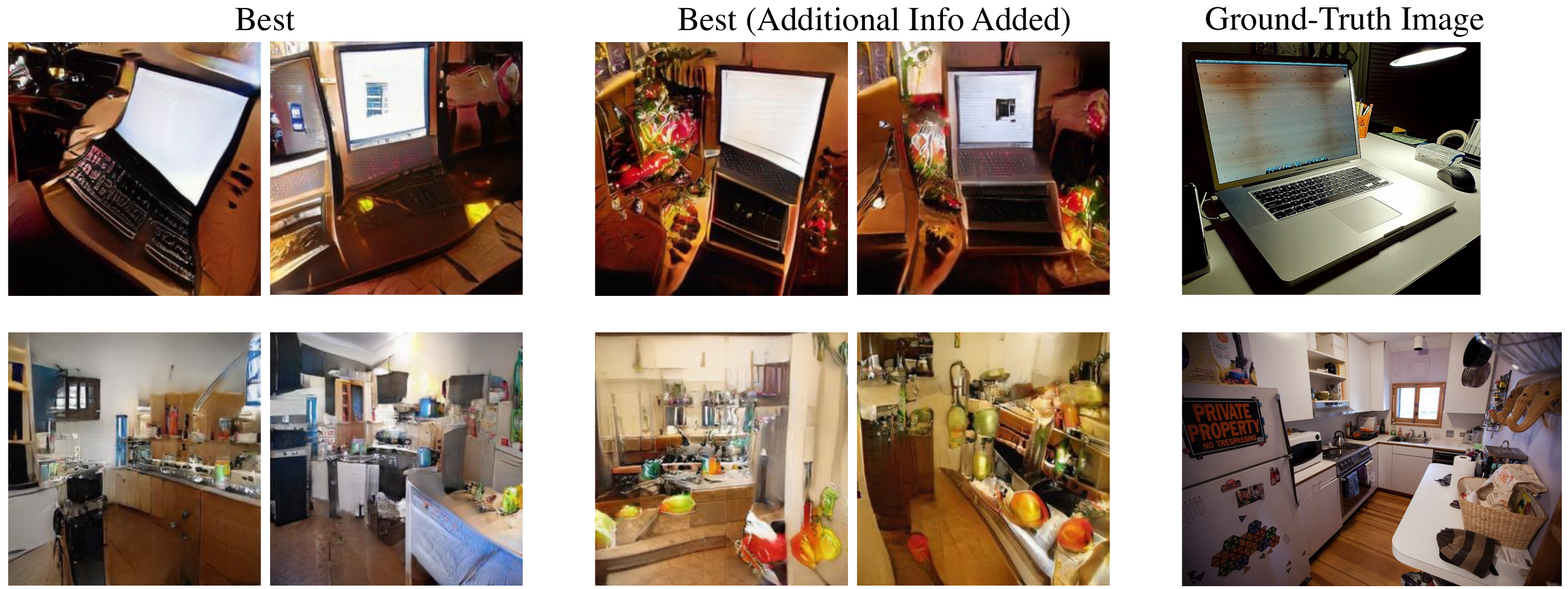}
\caption{Images generated with and without manually added information (COCO-CN Test set). The resolution of the generated images is $256\times256$ pixels; ground-truth images are shown in their original sizes respectively.}
\label{fig:addinfo}
\end{figure*}

To better understand what kind of information \mbox{\method} extracts from \textsc{en} translations, we also try to manually add additional information to \textsc{en} translations (the additional information does not appear in and is not added to the original $\mathbb{L}$ input). Of course, this section is for probing purposes only since MT systems are not likely to produce the same translations. We found that when the additional information is added to only several of the $12$ \textsc{en} translations, it can hardly get reflected in the generated image. Here, we show two COCO-CN test set examples in Figure~\ref{fig:addinfo} where we add the new information into $12$ \textsc{en} translations simultaneously. In its first and second rows, the original $\mathbb{L}$ input is `An open laptop is on the table.' and `It's a clean, but crowded kitchen.' respectively (translated from the original Chinese captions). We manually add new objects `roses' and `fruits' respectively to all their \textsc{en} translations as in Table~\ref{table:addinfo}. As seen in Figure~\ref{fig:addinfo}, the roses and fruits do appear in the generated images.

\begin{table*}[ht!]
\begin{center}
\resizebox{0.97\linewidth}{!}{%
\begin{tabular}{ll}
\toprule 
\rowcolor{Gray}
\multicolumn{1}{c}{\bf Original $\mathbb{L}$ Input}  &\multicolumn{1}{c}{\bf Modified \textsc{en} Translations}\\

\midrule

\multirow{12}{*}{\zh{桌子上摆放着一个打开的笔记本电脑。}} &\multicolumn{1}{c}{On a table is put on an open laptop\textcolor{myred}{\underline{, and roses}}.}\\
&\multicolumn{1}{c}{It was on the desk with an open laptop\textcolor{myred}{\underline{, and roses}}.}\\
&\multicolumn{1}{c}{There's a computer that's open that has an open laptop sitting on the table\textcolor{myred}{\underline{, and roses}}. }\\
&\multicolumn{1}{c}{There's a opened laptop on the table\textcolor{myred}{\underline{, and roses}}.}\\
&\multicolumn{1}{c}{There's an open laptop sitting on the table\textcolor{myred}{\underline{, and roses}}.}\\
&\multicolumn{1}{c}{An open laptop's on the table\textcolor{myred}{\underline{, and roses}}.}\\
&\multicolumn{1}{c}{And we have a laptop on your desk that's open\textcolor{myred}{\underline{, and roses}}.}\\
&\multicolumn{1}{c}{There was a laptop that was open on the table\textcolor{myred}{\underline{, and roses}}.}\\
&\multicolumn{1}{c}{A computer that opened up his laptop is in place on the table\textcolor{myred}{\underline{, and roses}}.}\\
&\multicolumn{1}{c}{There was a computer on the table. There was an open laptop on the table\textcolor{myred}{\underline{, and roses}}.}\\
&\multicolumn{1}{c}{There's an open laptop on the table\textcolor{myred}{\underline{, and roses}}.}\\
&\multicolumn{1}{c}{There was an unopened laptop on the table\textcolor{myred}{\underline{, and roses}}.}\\

\midrule
\multirow{12}{*}{\zh{这是一个干净，但拥挤的厨房。}} &\multicolumn{1}{c}{That's a clean-up, but crowded kitchen \textcolor{myred}{\underline{full of fruits}}.}\\
&\multicolumn{1}{c}{It's clean but crowded in the kitchen \textcolor{myred}{\underline{full of fruits}}.}\\
&\multicolumn{1}{c}{That's a clean, but crowded kitchen \textcolor{myred}{\underline{full of fruits}}.}\\
&\multicolumn{1}{c}{It's a clean, but crowded kitchen \textcolor{myred}{\underline{full of fruits}}.}\\
&\multicolumn{1}{c}{It's a clean-up but congested kitchen \textcolor{myred}{\underline{full of fruits}}.}\\
&\multicolumn{1}{c}{That's a clean, but congested kitchen \textcolor{myred}{\underline{full of fruits}}.}\\
&\multicolumn{1}{c}{And it's a clean, but crowded kitchen \textcolor{myred}{\underline{full of fruits}}.}\\
&\multicolumn{1}{c}{It's a clean, but congested kitchen \textcolor{myred}{\underline{full of fruits}}.}\\
&\multicolumn{1}{c}{- IT'S THIS IS A cleanING BUT CLOTHED CLIMBEN COILLOR IN THE CRUCKIT. - [CLICKS] \textcolor{myred}{\underline{full of fruits}}.}\\
&\multicolumn{1}{c}{It was a clean but crowd-cooked kitchen \textcolor{myred}{\underline{full of fruits}}.}\\
&\multicolumn{1}{c}{That's a clean one, but crowd-cooked kitchen \textcolor{myred}{\underline{full of fruits}}.}\\
&\multicolumn{1}{c}{It's a clean, but congested kitchen \textcolor{myred}{\underline{full of fruits}}.}\\

\bottomrule
\end{tabular}
}
\caption{Additional information added to the \textsc{en} translations. The underlined \textcolor{myred}{\underline{texts}} in red are added phrases. Removing the phrases derives the NMT-generated translations.}
\label{table:addinfo}
\end{center}
\end{table*}

\end{document}